\documentclass[a4paper,fleqn]{cas-dc}
\usepackage[numbers]{natbib}
\usepackage{multirow}
\usepackage{subcaption}
\usepackage{amsmath}
\usepackage{algpseudocode}
\usepackage{commath}
\usepackage{algorithm}
\usepackage{adjustbox}
\usepackage{graphicx}
\usepackage{textcomp}
\usepackage{adjustbox}
\usepackage{booktabs}
\def\tsc#1{\csdef{#1}{\textsc{\lowercase{#1}}\xspace}}
\tsc{WGM}
\tsc{QE}
\tsc{EP}
\tsc{PMS}
\tsc{BEC}
\tsc{DE}

\begin{document}
\let\WriteBookmarks\relax
\def\floatpagepagefraction{1}
\def\textpagefraction{.001}
\shorttitle{Fish Disease Detection Using Image Based Machine Learning Technique in Aquaculture}
\shortauthors{Md Shoaib Ahmed et~al.}

\title [mode = title]{Fish Disease Detection Using Image Based Machine Learning Technique in Aquaculture}                      

\author[1]{Md Shoaib Ahmed}[orcid=0000-0002-6938-7309]
 \cormark[1]
\fnmark[1]
\ead{shoaibmehrab011@gmail.com}


 \address[1]{Department of Computer Science and Engineering, Jahangirnagar University, Savar, Dhaka - 1342, Bangladesh }

\author[1]{Tanjim Taharat Aurpa}
 \fnmark[2]
 \ead{taurpa22@gmail.com}



\author
 [1]
{Md. Abul Kalam Azad}
 \fnmark[3]
 \ead{makazad@juniv.edu}


\cortext[cor1]{Corresponding author}
\begin{abstract}
Fish diseases in aquaculture constitute a significant hazard to nutriment security. Identification of infected fishes in aquaculture remains challenging to find out at the early stage due to the dearth of necessary infrastructure. The identification of infected fish timely is an obligatory step to thwart from spreading disease. In this work, we want to find out the salmon fish disease in aquaculture, as salmon aquaculture is the fastest-growing food production system globally, accounting for 70 percent (2.5 million tons) of the market. In the alliance of flawless image processing and machine learning mechanism, we identify the infected fishes caused by the various pathogen. This work divides into two portions. In the rudimentary portion, image pre-processing and segmentation have been applied to reduce noise and exaggerate the image, respectively. In the second portion, we extract the involved features to classify the diseases with the help of the Support Vector Machine (SVM) algorithm of machine learning with a kernel function. The processed images of the first portion have passed through this (SVM) model. Then we harmonize a comprehensive experiment with the proposed combination of techniques on the salmon fish image dataset used to examine the fish disease. We have conveyed this work on a novel dataset compromising with and without image augmentation. The results have bought a judgment of our applied SVM performs notably with 91.42 and 94.12 percent of accuracy, respectively, with and without augmentation.
\end{abstract}


\maketitle

\section{Introduction}

\begin{figure}
\centering
\includegraphics[width=1.0\columnwidth]{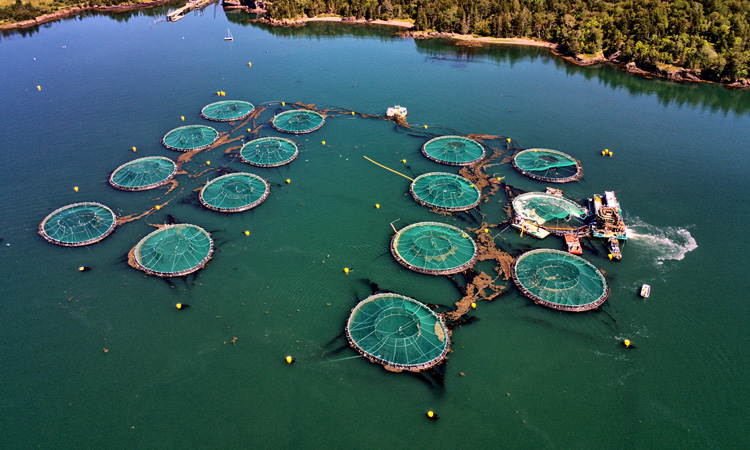}
\caption{Aquaculture~\protect\cite{FigAquaculture}}  
\label{fig:Aquaculture}
\end{figure}

The word aquaculture is related to firming, including breeding, raising, and harvesting fishes, aquatic plants, crustaceans, mollusks, and aquatic organisms. It involves the cultivation of both freshwater and saltwater creatures under a controlled condition and is used to produce food and commercial products as shown in Figure \ref{fig:Aquaculture}.
There are mainly two types of aquaculture. The first one is \textbf{Mariculture} which is the farming of marine organisms for food and other products such as pharmaceuticals, food additives, jewelry (e.g., cultured pearls), nutraceuticals, and cosmetics. Marine organisms are farmed either in the natural marine environment or in the land- or sea-based enclosures, such as cages, ponds, or raceways. Seaweeds, mollusks, shrimps, marine fish, and a wide range of other minor species such as sea cucumbers and sea horses are among the wide range of organisms presently farmed around the world's coastlines. It contributes to sustainable food production and the economic development of local communities. However, sometimes at a large scale of marine firming become a threat to marine and coastal environments like degradation of natural habitats, nutrients, and waste discharge, accidental release of alien organisms, the transmission of diseases to wild stocks, and displacement of local and indigenous communities \cite{MariCulture}.

The second one is \textbf{Fish farming} which is the cultivation of fish for commercial purposes in human-made tanks and other enclosures. Usually, some common types of fish like catfish, tilapia, salmon, carp, cod, and trout are firmed in these enclosures. Nowadays, the fish-farming industry has grown to meet the demand for fish products \cite{FishFarm}. This form of aquaculture is widespread for a long time as it is said to produce a cheap source of protein.

Global aquaculture is one of the quickest growing food productions, accounting for almost 53\%  of all fish and invertebrate production and 97\% of the total seaweed manufacture as of 2020. Estimated global production of farmed salmon stepped up by 7 percent in 2019, to just over 2.6 million tonnes of the market \cite{AquacultureIntroduction}. Global aquaculture of salmon has a threat of various diseases that can devastate the conventional production of salmon.

Diseases have a dangerous impact on fishes in both the natural environment and in aquaculture. Diseases are globally admitted as one of the most severe warnings to the economic success of aquaculture. Diseases of fishes are provoked by a spacious range of contagious organisms such as bacteria, viruses, protozoan, and metazoan parasites. Bacteria are accountable for the preponderance of the contagious diseases in confined fish \cite{FishDiseaseIntroduction}. Infectious diseases create one in every foremost vital threat to victorious aquaculture. The massive numbers of fishes gathered in a tiny region gives an ecosystem favorable for development and quickly spreads contagious diseases. In this jam-packed situation, a comparatively fabricated environment, fishes are stressed and also respond to disease. Furthermore, the water ecosystem and insufficient water flow make it easier for the spread of pathogens in gathered populations \cite{FishDiseaseIntroduction2}. Detection of disease with the cooperation of some image processing can help to extract good features.

Image segmentation becomes indispensable for various research fields like computer vision, artificial intelligence, etc. The \textit{k} means segmentation is a popular image processing technique that mainly partitions different regions in an image without loss of information. In \cite{kailasanathan2001image}, authors applied \textit{k} means segmentation for authentication of images. Another application of \textit{k} means segmentation shown at \cite{gaur2015handwritten} where they use this technique to recognize handwritten Hindi characters.

One of the most popular supervised machine learning techniques, support vector machine (SVM), has brought convenient solutions for many classification problems in various fields. It is a powerful classification tool that brings out quality predictions for unlabeled data. In \cite{khan2016analysis} Authors built an SVM model based on three kernel functions to differentiate dengue human infected blood sera and healthy sera. For image classification, another SVM architecture has been proposed in \cite{agarap2017architecture} where they emulate the architecture by combining convolutional neural network (CNN) with SVM. SVM provides remarkable accuracy in many contexts. 

In this paper, we conduct our research on the salmon fish disease classification, either the fish has an infection or not, with a machine vision-based technique. A feature set is a trade-off for the classification of the disease. Image processing techniques are used to extort the features from the images, then a support vector machine (SVM) is employed for the successful classification of infectious disease. Hither, we summarize the entire concept of this work’s contribution given below:

\begin{itemize}
    \item Propose a groundbreaking framework for fish disease detection based on the machine learning model (SVM).
    
    \item Appraising and analyzing the performance of our proposed model both with and without image augmentation.
    
    \item Juxtaposing our proposed model with a good performing model by some evaluation metrics.
    
\end{itemize}

\section{Related Work}

Some works focused on only some basic image processing techniques for the identification of fish disease. Shaveta et al. \cite{LRShaveta} proposed an image-based detection technique where firstly applies image segmentation as an edge detection with Canny, Prewitt, and Sobel. However, they did not specify the exact technique that engrossed for feature extraction. In feature extraction, they applied Histogram of Gradient (HOG) and Features from Accelerated Segment Test (FAST) for classification with a combination of both techniques. They tried to discover a better classification with a combination instead of applying a specific method with less exactness. Another technique Lyubchenko et al. \cite{LRLyubchenko} proposed a structure called the clustering of objects in the image that obliged diverse image segmentation actions based on a scale of various clusters. Here, they chose markers for individual objects and objects encountered with a specific marker. Finally, they calculated the proportion of an object in the image and the proportion of infected area to the fish body to identify fish disease. However, individual marking of an object is time-consuming and not effective.

There are some approaches focused on the consolidation of image processing and machine learning. Malik et al. \cite{LRMalik} proposed a specific fish disease called Epizootic Ulcerative Syndrome (EUS) detection approach. Aphanomyces invadans, a fungal pathogen, cause this disease. Here, they approached combination styles that combine the Principal Component Analysis (PCA) and Histogram of Oriented Gradients (HOG ) with Features from Accelerated Segment Test (FAST) feature detector and then classify over machine learning algorithm (neural network). The sequence of FAST-PCA-NN gives 86 percent accuracy through the classifier, and HOG-PCA-NN gives 65.8 percent accuracy that is less than the previous combination.

Verma et al. \cite{verma2017analysis} proposed a sensitive topic that is kidney stone detection. In this paper, the authors apply morphological operations and segmentation to determine ROI (region of interest) for the SVM classification technique. After applying this technique, they investigated the kidney stone images with some difficulties, such as the similarity of kidney stone and low image resolution. Zhou et al. \cite{zhou2017device} introduced a device-free present detection and localization with SVM aid. Here, the detection algorithm can detect human presence through the SVM classifier using CSI (channel state information) fingerprint. Trojans in hardware detection \cite{inoue2017designing} depend on SVM based approach. Here, the authors evaluated a trojans detection method with their designed hardware. For SVM analysis, their netlists consist of three types of hardware trojan with normal and abnormal behavior. 

We can conclude that none has performed any depth research work on salmon fish disease classification regarding the research obligations described above. Furthermore, most of the research works involved typical fish disease classification but not in aquaculture. All those described techniques depend solely on image processing or a combination of image processing and machine learning technique but not up to the mark.


\section{Preliminary and Proposed Framework}
\begin{figure*}
 \begin{center}
	\centering
	\includegraphics[scale=.90]{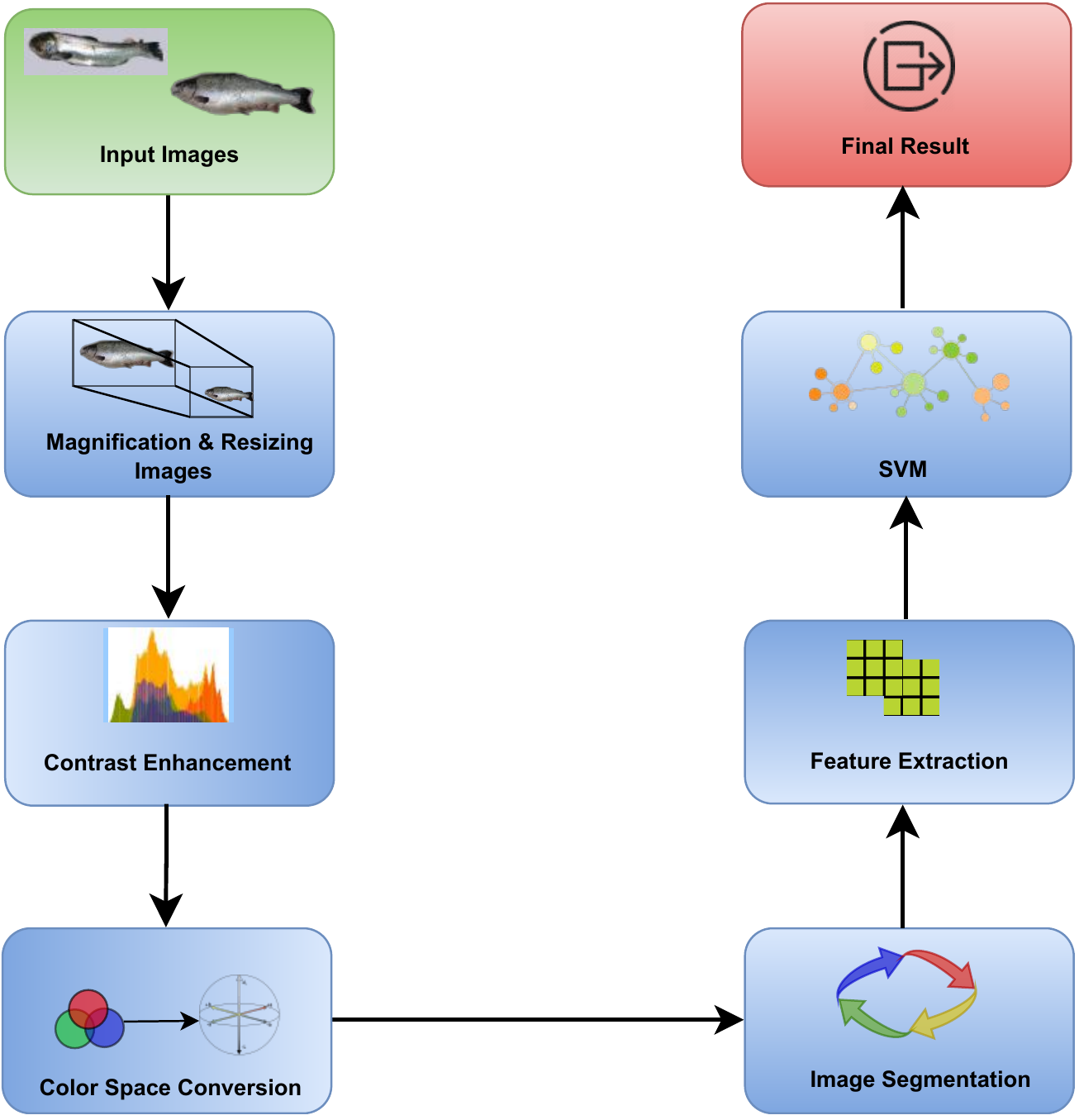}

 	\caption{Proposed Framework (The overall anatomy of our proposed work gradually from input to result).}
	\label{img:proposedFramework}
	\end{center}
 \end{figure*}

This section has several stages presented in Figure \ref{img:proposedFramework}. Here we precisely present the appurtenant technologies and a solution framework of salmon fish disease classification.

\subsection{Cubic Splines Interpolation}
Raw images appeared on the dataset in various sizes. If we do not resize these images before training the classifier, the classifier's efficiency may be decreased. As we collected these images from different sources, we reshape them before applying them to the classifier.

For image magnification and fixed-size conversion, we use an improved interpolation method called extended \textbf{\textit{Cubic Splines interpolation}} \cite{BSPline}. For a finite interval $[a,b]$, let $\left \{ x_{i} \right \}_{i = 0}^{n}$ be a partition of the interim with steady step size, $h$. We elongate the partition using Equation \ref{eqn:cubicB1}.

\begin{equation}\label{eqn:cubicB1}
h = \frac{b - a}{n}, \quad x_{0} = a, \quad x_{i} = x_0 + ih, \quad i = \pm 1, \, \pm 2, \, \pm 3, ...
\end{equation}

Given $\left \{ x_{i} \right \}$, the extended cubic B-spline function, $S\left ( x \right )$ is a linear combination of the extended cubic B-spline basis function in Equation \ref{eqn:cubicB2}, 

\begin{equation}\label{eqn:cubicB2}
S\left ( x \right ) = \sum_{i = -3}^{n-1} C_{i}EB_{3,i}\left ( x \right ), \quad x \, \epsilon \, \left [ x_{0}, x_{n} \right ]
\end{equation}

where $C_{i}$ are unknown real coefficients. Since  $\,C_{i}EB_{3,i}\left ( x \right )$ has a support on $\left [ x_{i}, x_{i+4} \right ]$, there are three nonzero basis function evaluated at each $x_{i}$: $C_{i}EB_{3,i-3}\left ( x \right )$, $C_{i}EB_{3,i-2}\left ( x \right )$, $C_{i}EB_{3,i-1}\left ( x \right )$.

\subsection{Adaptive Histogram Equalization}
In order to improve the image's quality, contrast enhancement is an essential technique. It contributes to recover the lost information in images. Due to magnification and resizing images, some images may lose information. To prevent this problem, we use Adaptive Histogram Equalization and enhance the contrast of each image.

Adaptive histogram equalization (AHE) is a vision processing approach used to enhance contrast in images. Here, we use an extension of AHE called contrast limited adaptive histogram equalization (CLAHE) \cite{liu2019adaptive}. CLAHE diverges from conventional AHE in its contrast limiting. A user-defined value called clip limit where CLAHE limits the augmentation by clipping the histogram. The amount of clamor in the histogram depends on the clipping level. Also, the smoothness and enhancement of contrast rely on this clipping level too. A modification of the limited contrast technique called adaptive histogram clip (AHC) can also be applied. AHC dynamically calibrates the clipping level and balanced over-enhancement of a background area of images \cite{hitam2013mixture}. Here, we use one of the AHC called the Rayleigh distribution that forms a normal histogram. Equation \ref{eqn:clahe} represents this function as below

\begin{equation}\label{eqn:clahe}
Rayleigh \;\;\; p = p_{min} + [\;2(\alpha^2)\;\;ln(\frac{1}{1-Q(f)})\;]^{0.5}
\end{equation}

Where $p_{min}$ and $Q(f)$ represent minimum pixel value and cumulative probability distribution, respectively. $\alpha$ is a non-negative real scalar indicating a distribution parameter. In this experiment, we set 0.01 as a clip limit and 0.04 as the $\alpha$ value in the Rayleigh distribution function.

\subsection{RGB Color Space to L*a*b Color Space}
We now convert the adaptive histogram equalized image from RGB to L*a*b. Hence, we use the \textit{k}-means clustering for segmentation of the image, and here \textit{k}-means clustering technique segments image efficiently in L*a*b color space rather than RGB color space \cite{burney2014k}. In L*a*b color space, where L expresses the lightness of an image and the a, b color channel depicts the other color combinations \cite{rahman2016non}. For this transformation, we need to convert from RGB color space to XYZ color space \cite{ColorCoversion}, \cite{bianco2007new} according to Equation \ref{eqn:rgbToXyz}.
\begin{equation}\label{eqn:rgbToXyz}
\begin{bmatrix}
X\\ 
Y\\ 
Z
\end{bmatrix} = \begin{bmatrix}
0.412453 & 0.357580 & 0.180423\\ 
0.212671 &  0.715160  & 0.072169  \\ 
0.019334 & 0.119193   & 0.950227
\end{bmatrix}
*
\begin{bmatrix}
R\\ 
G\\ 
B
\end{bmatrix}
\end{equation}

Now, transforming XYZ color space to L*a*b color space \cite{acharya2002median} according to equation [\ref{eqn:xyzToLab} - \ref{eqn:lab_a}] and suppose that the tristimulus values of the reference white are $X_n$, $Y_n$, $Z_n$.

\begin{equation}\label{eqn:xyzToLab}
L^* = \left\{\begin{matrix}
116(\frac{Y}{Y_n})^\frac{1}{3} - 16 & if \frac{Y}{Y_n} > 0.008856 \\ \\
903.3 \frac{Y}{Y_n} & if  \frac{Y}{Y_n} \leq  0.008856
\end{matrix}\right.
\end{equation}
\begin{equation}\label{eqn:lab_a}
a^* = 500 (f(\frac{X}{X_n})- f(\frac{Y}{Y_n}))
\end{equation}
\begin{equation}\label{eqn:lab_b}
b^* = 200 (f(\frac{Y}{Y_n})- f(\frac{Z}{Z_n}))
\end{equation}

Where, 
\begin{equation}\label{eqn:lab_a}
f(t) = \left\{\begin{matrix}
t^\frac{1}{3} & if \; t > 0.008856 \\ \\
7.787t +  \frac{16}{116} & if  \; t \leq  0.008856
\end{matrix}\right.
\end{equation}

\subsection{\textit{k}-means Clustering Segmentation}
When we segment the infected part in a fish's image, the classifier learns to identify the infected fish's images accurately. In this subsection, the converted image is then segmented, utilizing the \textit{k}-means clustering technique into several regions. As a result, it separates infected areas from fishes' images. Techniques in \cite{gaur2015handwritten} and \cite{hartigan1979algorithm} adheres to the conventional steps to appease the primary intent, and that is clustering the image objects into K distinct groups. The steps of \textit{k}-means clustering technique as follows.
\begin{enumerate}
    \item Determine the total number of clusters k.
    \item In each group, choose k points as a centroid.
    \item Appoint each data point to the nearest centroid that assembles k clusters.
    \item Calculate and assign the new centroid of each cluster.
   \item Go to step 4 when any reassignment took place, and that time reassigns each data point to the nearest centroid. Otherwise, the model is ready.
\end{enumerate}

The function of the \textit{k}-means clustering technique \cite{de2009detection} is the minimum square structure that is measured by
\begin{equation}\label{eqn:KMeans}
J = \sum_{j = 1}^{k}\sum_{i=1}^{n}\left \| {x_i}^{(j)} - c_j \right \|^2
\end{equation} 

Where $J$ is the objective function that displays the similarity act of the $n$ objects encompassed in specific groups. $k$ and $n$ are the numbers of clusters and number of cases, respectively. Finally, $ \left \| {x_i}^{(j)} - c_j \right \| $ is the distance function from a point ${x_i}^{(j)}$ to the centroid of group $c_j$.

Here, two types of feature vectors are acquired from the infected area in a fish, i.e., co-occurrence and statistical. We are going to explain certain features with nicety in the experimental evaluation section.

\subsection{Support Vector Machine}
\begin{figure*}
 \begin{center}
	\centering
	\includegraphics[scale=.85]{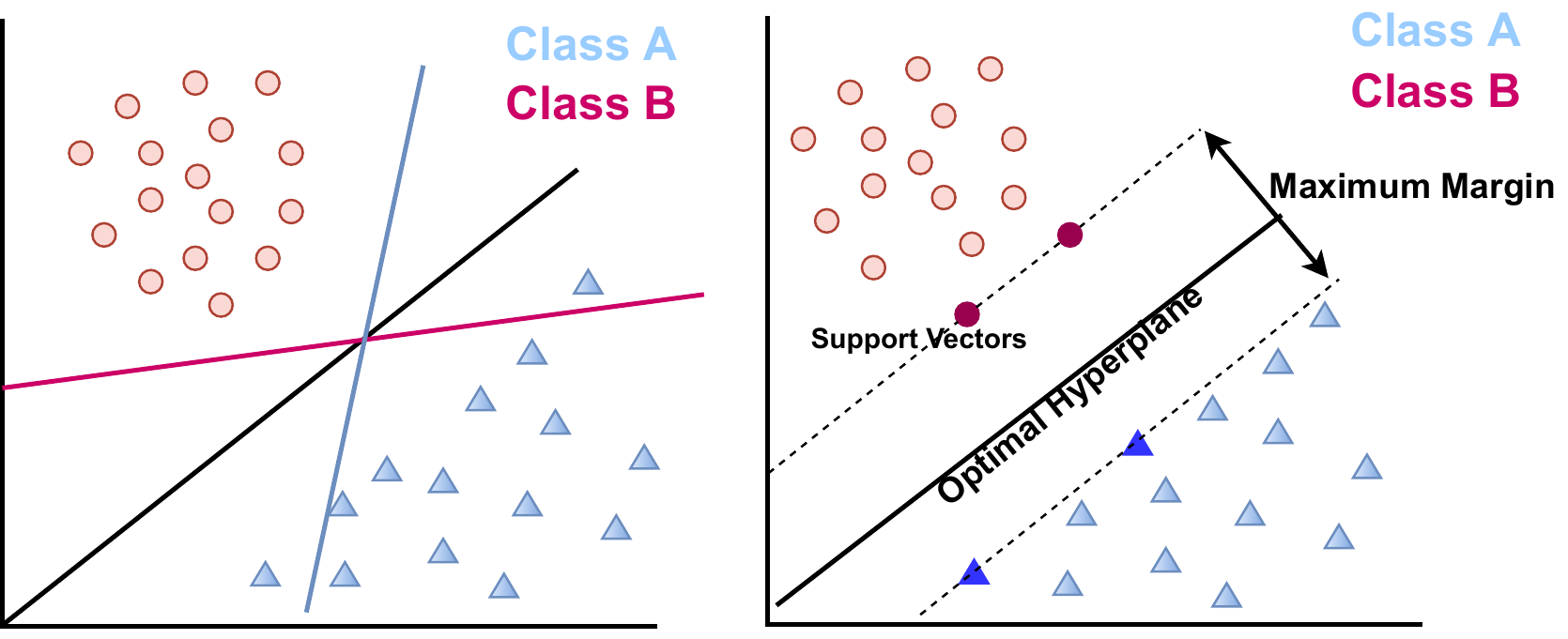}

 	\caption{Support Vector Machine (Discovering the optimal hyperplane and the separation of classes for optimal hyperplane).}
	\label{img:svm}
	\end{center}
 \end{figure*}

We exploit the feature vectors discussed in the previous subsection to SVM. Support vector machine (SVM) is a supervised machine learning algorithm used in many classification problems for its higher accuracy rate. It aims to construct a hyperplane between different classes with a margin to classify objects. The hyperplane can be constructed in a multidimensional axis to partitioned the data points \cite{meyer2003support, noble2006support}. Figure \ref{img:svm} indicates the basic diagram for the support vector machine.  Some common terms related to SVM are mentioned below:

\textbf{Optimal Hyperplane:} The boundary that distinguishes two classes with the maximum margin is the optimal hyperplane. It is an N-1 dimensional subset of N-dimensional surface that distinguishes the classes on that surface. In two dimensions, the hyperplane is a line. With the increasing number of dimensions, the hyperplane's dimension is increased. The optimal hyperplane is determined as $wx_i + b =0$. Here $w$ is the weight vector, $x$ is the input feature vector, and $b$ is the bias. For all aspects of the training set, the w and b comply with the following inequalities \cite{suthaharan2016support}:
\begin{center}
$wx_i + b \geq +1 \; \;if \; y_i = 1$\\
$wx_i + b \leq -1 \; \;if \; y_i = -1$
\end{center}

\textbf{Support Vectors:} Data points that are more adjacent to the hyperplane and influence the positioning of the hyperplane are known as support vectors. The more similar points between the two classes become the support vectors. These points avail in the establishment of SVM. Suppose a labeled training dataset represented as ${\{(x_i,y_i) \;|\; i = 1,2,...,k\}}$, Where $x_i$  is a feature vector representation or input, and $y_i$ is the class label or output. To consider the margin among two kinds of points to maximize, use the Lagrange technique to revamp the original problem to identify a maximum value of the function by Equation \ref{eqn:SVM_1}.

\begin{equation}\label{eqn:SVM_1}
Q(\alpha ) = \sum_{i=1}^{k}\alpha _i - \sum_{i,j=1}^{k}\alpha _i \alpha_j y_i y_j(x_i.y_j)
\end{equation}
Where $\alpha _i$  is the reciprocal multiplier of every snippet. Then remodel it to a large dimensionality space by kernel function $K(x_i,y_j)$ shown as Equation \ref{eqn:SVM_2}.
\begin{equation}\label{eqn:SVM_2}
Q(\alpha ) = \sum_{i=1}^{k}\alpha _i - \sum_{i,j=1}^{k}\alpha _i \alpha_j y_i y_jK(x_i.y_j)
\end{equation}

\textbf{Margin:} Margin is the gap between two non-overlapping classes separated by hyperplanes. It mainly indicates the gap between data points and the dividing line. For the optimal hyperplane, we required the maximum margin.

\textbf{Kernel:} The functions used by the SVM algorithm to classify the objects are mainly known as the kernel function. They mainly transform the inputs into a required form to construct the hyperplane easily. There are many kernels \cite{zhang2015complete} used in SVM such as linear, polynomial (homogeneous and heterogeneous), gaussian, fisher, graph, string, tree, etc. 

The \textbf{linear kernel} is one of the most used and straightforward kernel functions used for the linearly separable data points \cite{ben2010user}.

There are many application areas where SVM usually outperforms any other classifier with high accuracy \cite{chandra2018survey}. It is mainly designed for the binary classification problem that we have addressed here. The SVM is trained using feature training datasets for robust performance in the test dataset. For performance analysis, we need to evaluate some metrics shown in the experimental evaluation section.

\subsection{System Architecture}
\begin{figure*}
	\centering
	\includegraphics[scale=.65]{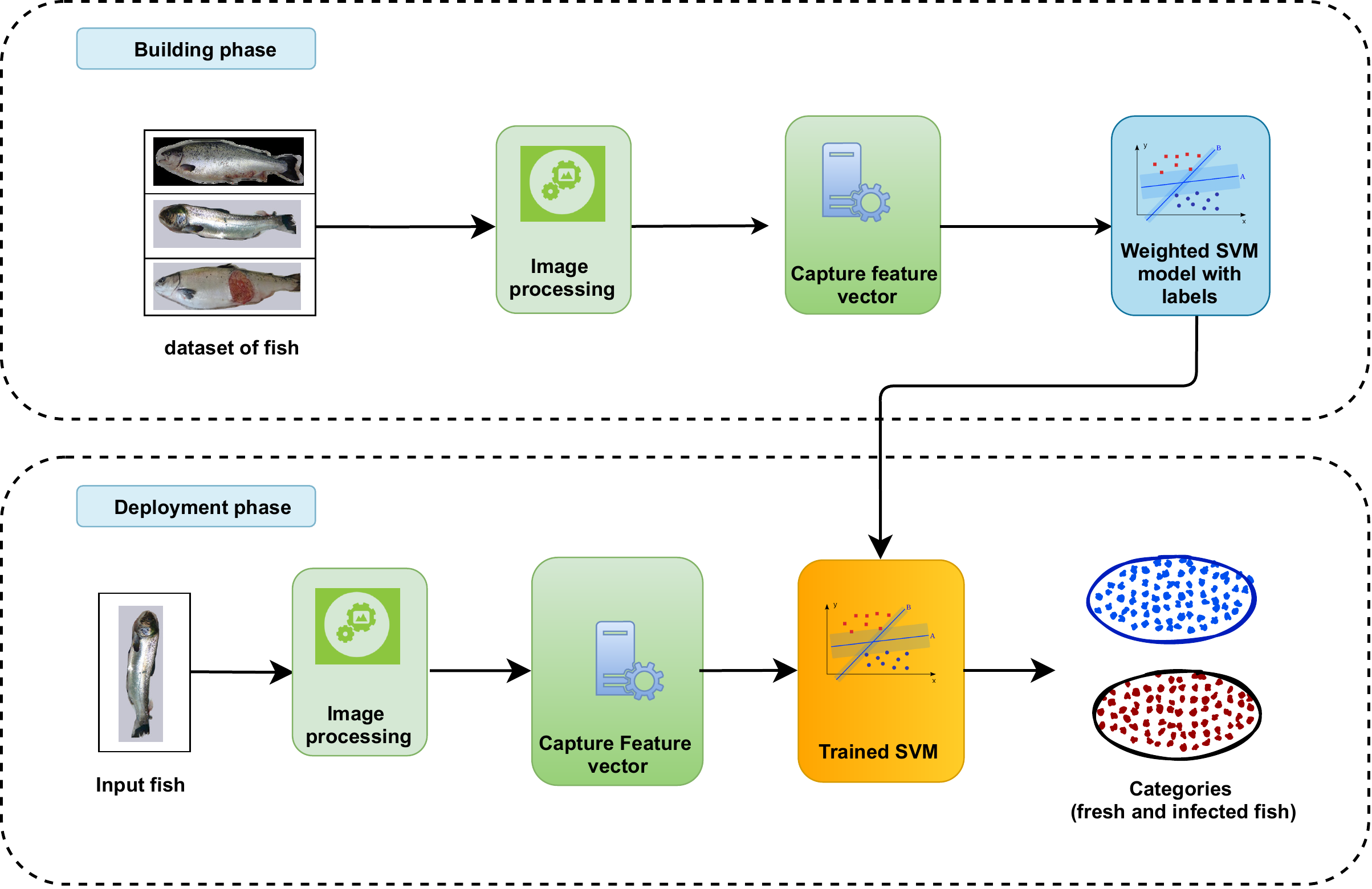}

 	\caption{System Architecture (A well-regulated diagram demonstrating the entire process from data acquisition to model training and prediction of classes).}
	\label{Fig:SystemDiagram}
 \end{figure*}

We embellish a system architecture shown in Figure \ref{Fig:SystemDiagram}. It contains two phases, the first one is the building phase, and the second one is the deployment phase. Inside the building phase, we process the labeled images as training data.
\begin{itemize}
    \item Each image is refined through the serialization of mentioned image processing techniques such as cubic splines interpolation, adaptive histogram equalization, and covert RGB color space to L*a*b color space.
    \item \textit{K}-means clustering technique is applied for the image segmentation and identifies two types of feature vectors, namely co-occurrence matrix features and statistical features, respectively.
    \item Exploit these feature vectors to SVM for further processing.
\end{itemize}

The weighted model is used to train the feature vectors and the corresponding labels in the building phase.

The outgrowth from the building phase comprises a trained or learned SVM model. This trained model is applied for classifying any incoming fish in the deployment phase. In the deployment phase, some steps are performing as follows.
\begin{itemize}
    \item An input of any fish image is supplied into the system, then the image is refined through the serialization of image processing techniques.
    \item Apprehended two types of feature vectors by the \textit{K}-means clustering technique in terms of image segmentation.
    \item From the feature extractor, the feature vectors are fed for the trained SVM model.
    \item Finally, the outcome is a label of an input image to classify the specific category as fresh or infected fish.
\end{itemize}

This system architecture exhibits the entire process from data acquiring to model training and prediction of classes.

\section{Evaluation}
This section delineates the experimental intervention in detail to evaluate our proposed approach. In this evaluation, we extract the features with a statistical and grey-level co-occurrence matrix (GLCM) with appropriate analogous terms utilized in our fish image dataset. For classification results, we need some performance evaluation metrics to show the operating capability to predict new data.

\subsection{Environment Specifications}
Here, we use the combination of MATLAB \footnote{https://www.mathworks.com/solutions/image-video-processing.html}  as a multi-paradigm programming language and Python. For image processing tasks such as cubic splines interpolation, adaptive histogram equalization, and image conversion from RGB to L*a*b, we have used MATLAB. Then, features extraction and training of our SVM model are employed by Python in the Google Colab \footnote{https://colab.research.google.com/} platform. Image interpretation and classification need enormous computing power. Hence, powerful computing tools installation with additional hardware support is expensive. So we use the Google Colab platform that serves us high-end CPU and GPU on the cloud premises that accommodate to train our model efficiently with less time. There is no extra encumbrance to install necessary packages because this platform crafts with all the obligatory packages used in the training process \cite{bisong2019google}. Google Colab ships NVIDIA K80 with 12 GB of GPU memory and 358 GB of disk space. This embellished environment gives an immense computational power to train machine learning models. 

\subsection{Experimental Dataset}
\begin{figure*}
    \centering
     \begin{subfigure}{.45\linewidth}
       \centering\includegraphics[scale=.17]{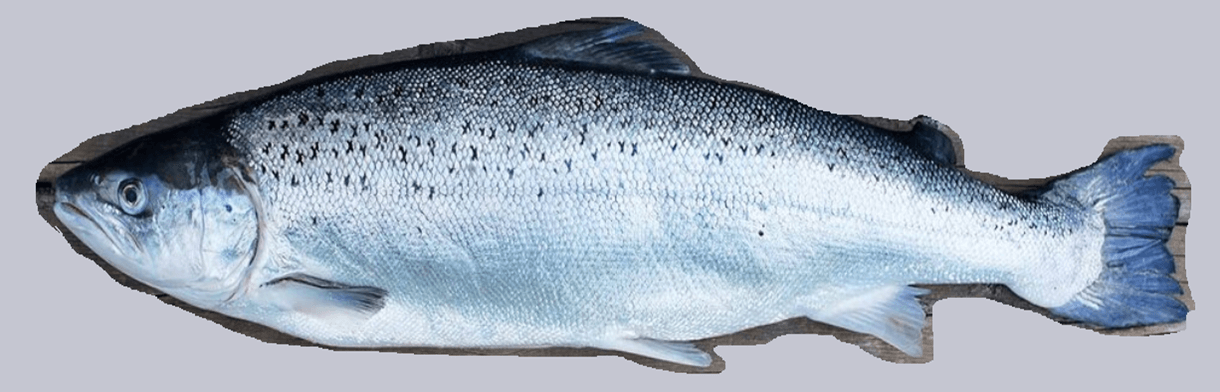}
       
     \end{subfigure}
     \begin{subfigure}{.45\linewidth}
       \centering\includegraphics[scale=.17]{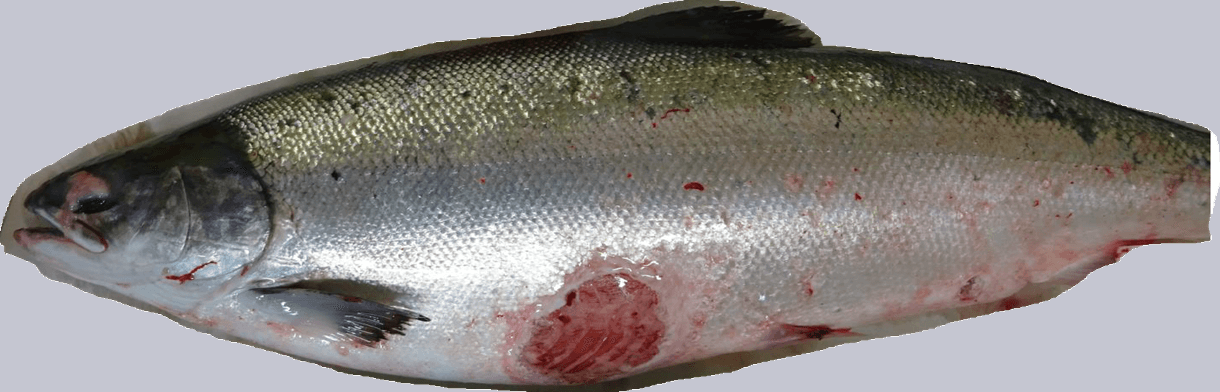}
       
     \end{subfigure}
     
    \medskip
     \begin{subfigure}{.45\linewidth}
       \centering\includegraphics[scale=.17]{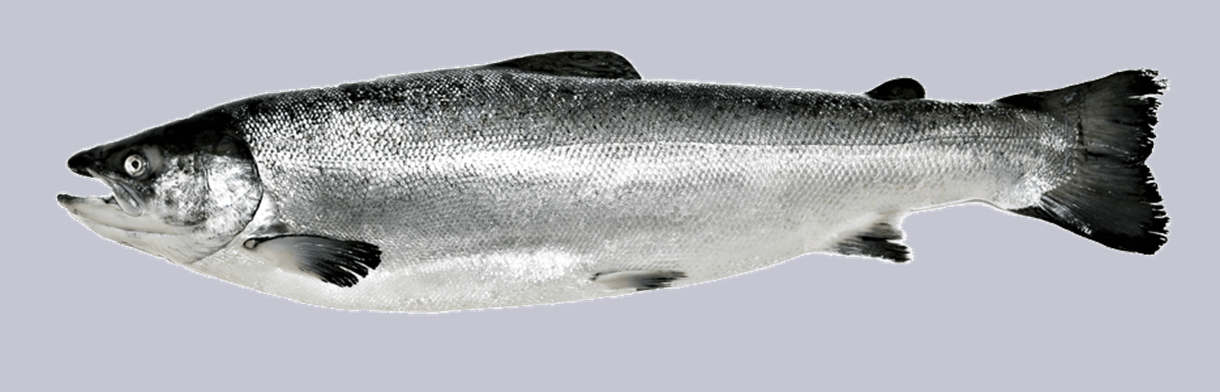}
       \caption{Fresh Fish}
     \end{subfigure}
     \begin{subfigure}{.45\linewidth}
       \centering\includegraphics[scale=.17]{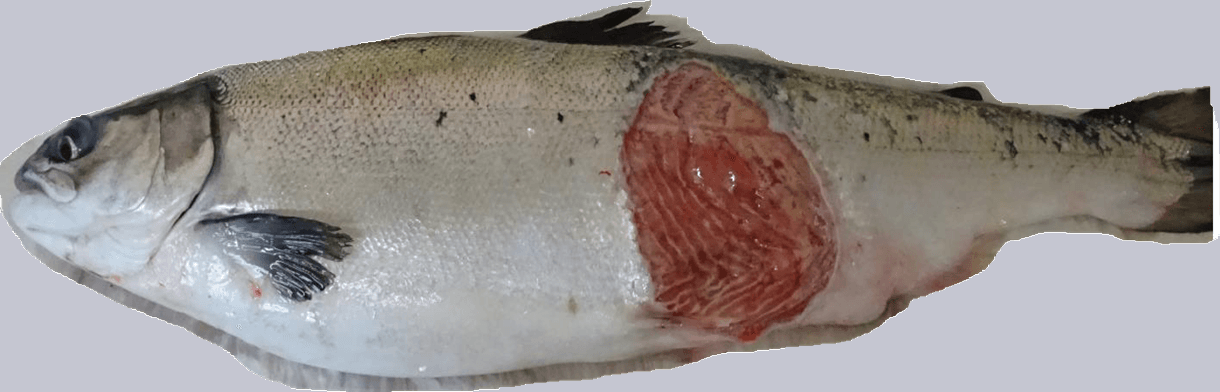}
       \caption{Infected Fish}
     \end{subfigure}
    
    \medskip

    \caption{Salmon Fish (Two sample of fresh fish and infected fish from our dataset).}
    \label{FIG:salmonFishExample}
\end{figure*}

Since there is no accessible dataset of salmon fresh and infected fish, we prepared a shibboleth and novel dataset - some images from the internet and most of them from some aquaculture firms. The dataset contains images of fresh and infected salmon fish displays in Figure \ref{FIG:salmonFishExample}. We collect a total of 266 images that are used to train and validate our model. For training and testing, we split our dataset with a ratio of train and test data that depicts in Table \ref{tab:datasetSplitting}. The total number of training and testing images are 231 and 35, respectively.

Hence, our data acquisition is a complicated process. We apply the image augmentation technique in our dataset for expanding the dataset. Here we use the \textit{image\textunderscore argumentor \begin{NoHyper}\footnote{https://github.com/codebox/image\textunderscore augmentor}\end{NoHyper}} tool with some image augmentation operations such as Horizontal Fip (fliph), Vertical Flip (flipv), Rotates (rot), Pixel Shifting (trans), and Zooms (zoom). After the augmentation operation, we perceive 1,105 training images depicted in Table \ref{tab:datasetSplittingWithAugmentaion}. The total number of training and testing images is 1,105 and 221, respectively. 

\begin{table}[]
\centering
\caption{Overall dataset splitting (The total number of fresh and infected fish images without augmentation).}
\label{tab:datasetSplitting}
\begin{tabular}{ccc}
\hline
\textbf{Fish}  & \textbf{Training images}  & \textbf{Testing images} \\ \hline
Fresh fish     &  68                       &  15                      \\ \hline
Infected fish  &  163                      &  20                      \\ \hline
\textbf{Total} & \textbf{231}              & \textbf{35}              \\ \hline
\bottomrule
\end{tabular}
\end{table}

\begin{table}[]
\centering
\caption{Overall dataset splitting (The total number of fresh and infected fish images with augmentation). }\label{tab:datasetSplittingWithAugmentaion}
\begin{tabular}{ccc}
\hline
\textbf{Fish}  & \textbf{Training images} & \textbf{Testing images} \\ \hline
Fresh fish     & 320                      & 64                      \\ \hline
Infected fish  & 785                      & 157                    \\ \hline
\textbf{Total} & \textbf{1,105}           & \textbf{221}             \\ \hline \bottomrule
\end{tabular}
\end{table}

\subsection{Features Extraction}

We consider two types of feature extraction techniques: one is statistical features, and another one is grey-level co-occurrence matrix (GLCM) features based on interpreting fish diseases. Statistical features are described as follows.
\begin{itemize}
    \item Mean ($\mu$): Presume there are $P$ pixels in infected regions and $Q$ pixels in fresh regions, and the gray-scale color intensity of a pixel in infected regions is $\psi$, then the mean $\mu$ is represented as the following Equation \ref{eqn:Fea_1}.
    
    \begin{equation}\label{eqn:Fea_1}
    \mu = \frac{\sum_{i=1}^{P}\psi_i}{P}
    \end{equation}
    
    \item Standard deviation ($\sigma$): Presume there are $P$ pixels in infected regions where the gray-scale color intensity of a pixel represents $\psi$ and mean gray-scale color intensity of all pixels represents $\mu$. The standard deviation $\sigma$ is defined as the following Equation \ref{eqn:Fea_2}.
    
    \begin{equation}\label{eqn:Fea_2}
    \sigma = \sqrt{\frac{\sum_{i=1}^{P}(\psi_i-\mu)^2}{P}} 
    \end{equation}
    
    \item Variance (${\sigma}^2$): If there are $P$ pixels in infected regions where the gray-scale color intensity of a pixel and mean gray-scale color intensity of all pixels represent $\mu$ and $\psi$, then the variance ${\sigma}^2$ is defined as the following Equation \ref{eqn:Fea_3}.
    
      \begin{equation}\label{eqn:Fea_3}
    {\sigma}^2 = {\frac{\sum_{i=1}^{P}(\psi_i-\mu)^2}{P}} 
    \end{equation}
    
    \item Kurtosis ($\kappa$): Presume there are $P$ pixels in infected regions where   $\psi$ and $\mu$ represent the gray-scale color intensity of a pixel and the mean gray-scale color intensity of all pixels respectively, then the kurtosis $\kappa$ is defined as the following Equation \ref{eqn:Fea_4}.
    
    \begin{equation}\label{eqn:Fea_4}
    \kappa  = \frac{\frac{1}{P}\sum_{i=1}^{P}(\psi-\mu)^4}{(\frac{1}{P}\sum_{i=1}^{P}(\psi-\mu)^2)^2}-3
    \end{equation}
    
    \item Skewness ($\gamma$):Here, the mean is $\mu$, the standard deviation is $\sigma$, and the mode is $\rho_m$ for the gray-scale color intensity of all pixels in infected areas. Then the skewness $\gamma$ is defined as the following Equation. \ref{eqn:Fea_5}.
    
     \begin{equation}\label{eqn:Fea_5}
    \gamma = \frac{\mu - \rho_m}{\sigma}
    \end{equation}
    
\end{itemize}

Along with those statistical features, an amount of GLCM features is used here. These are convenient for extracting textural features from images. 
By examining the relationship between two pixels at a time, an assessment of the intense diversity at the pixel can be executed.

Let us presume that $f(a,b)$ be a digital image of two dimensions with $X \times Y$ pixels and gray levels $G_L$. We also presume that $(a_1,b_1)$ and $(a_2,b_2)$ are two pixels in $f(a,b)$, the distance is $D$ and the angle between the dimension and the ordinate is $\theta$. So, a GLCM $M(i, j, D, \theta)$ depicts as the following equation \ref{eqn:glcm_1}.

 \begin{equation}\label{eqn:glcm_1}
    M(i, j, D, \theta) = \left | \{(a_1,b_1),(a_2,b_2) \epsilon X \times Y: \linebreak D,\theta,f(a_1,b_1) \} \right |
\end{equation}

In this experiment, we use five GLCM features, namely Contrast ($C$), correlation ($\chi$), energy ($\zeta$), entropy ($\Delta$), and homogeneity ($\xi$). Those are represented as the following equation of \ref{eqn:glcm_2} to \ref{eqn:glcm_6}.   
 \begin{equation}\label{eqn:glcm_2}
    Contrast \; \; C = \sum_{i=0}^{G_L-1}\sum_{j=0}^{G_L-1}(i-j)^2M(i,j)
\end{equation}
 \begin{equation}\label{eqn:glcm_3}
    Correlation \; \; \chi = \frac{\sum_{i=0}^{G_L-1}\sum_{j=0}^{G_L-1}i.j.M(i,j)-\mu_a.\mu_b}{\sigma_a.\sigma_b} 
\end{equation}
 \begin{equation}\label{eqn:glcm_4}
   Energy \; \; \zeta = \sum_{i=0}^{G_L-1}\sum_{j=0}^{G_L-1}M(i,j)^2
\end{equation}
 \begin{equation}\label{eqn:glcm_5}
  Entropy \; \; \Delta = -\sum_{i=0}^{G_L-1}\sum_{j=0}^{G_L-1}M(i,j)\log M(i,j) 
\end{equation}
 \begin{equation}\label{eqn:glcm_6}
  Homogeneity \; \; \xi = \sum_{i=0}^{G_L-1}\sum_{j=0}^{G_L-1}\frac{M(i,j)}{1+(i-j)^2}
\end{equation}

Here, $\mu_a$, $\mu_b$, $\sigma_a$, and $\sigma_b$ are the sum of anticipated and variance values individually for the row and column entries.

\subsection{Proposed Classifier}
Here, we exploit linear SVM for nonseparable sheaths as our classifier. Since the total number of training and testing images (without augmentation) are  231 and 35, respectively, the training dataset is $\{(x_1,y_1), (x_2, y_2), ...,(x_{231}, y_{231})\}$, where $x_i = (\mu, \sigma, {\sigma}^2,\kappa, \gamma, C, \chi, \zeta, \Delta, \xi)$ is the input vector and $y_i = \pm 1$. The reciprocal multiplier of $\{\alpha_1, \alpha_2,...,\alpha_{231}\}$ depicts as the following equation of \ref{eqn:SVM_Evaluation}.

\begin{equation}\label{eqn:SVM_Evaluation}
Q(\alpha ) = \sum_{i=1}^{35}\alpha _i - \frac{1}{2} \sum_{i,j=1}^{35}\alpha _i \alpha_j y_i y_j x_i x_j
\end{equation}
subject to the constraints\\

1. $\sum_{i=1}^{35}\alpha_iy_i = 0$\\ 

2. $0 \leq \alpha_i \leq C \; \; for \; i = 1,2,...,35$

\vspace{.5cm}
Where $C$ is a non-negative parameter functioning as an upper-bound value of $\alpha_i$.

The $C$ parameter is acknowledged as the penalty parameter. The $C$ parameter affixes a penalty for all misclassified data points. The low value of $C$ identifies the low points of misclassification. As a result, a large-margin decision boundary is taken at the expense of a higher number of misclassifications. Whereas the significant value of $C$, SVM seeks to reduce the number of misclassification resulting in a smaller margin decision boundary.

Here we settle all SVM parameters through a training process and apply a significant numeric value for $C$. We have used linear kernel in our work from our applied four kernels, namely linear, sigmoid, polynomial, and Gaussian. From those, we have seen that the accuracy varies a negligible amount, and the linear kernel has performed satisfactorily with a short time of the process.
\subsection{Performance Evaluation Metrics}

We appraise the appearance of our trained SVM model based on several metrics. For predicting new data or images, we use a confusion matrix to envision our model's performance. A confusion matrix comprises four building blocks, namely True Positive (TP), True Negative (TN), False Positive (FP), and False Negative (FN). TP and TN refer to the cases where the predictions are true and negative. FP refers to the positively false predictions, and FN indicates negatively false predictions \cite{marom2010using}. We compute more distinct metrics to evaluate our model from the confusion matrix. These precise metrics are Accuracy, Precision, Recall or Sensitivity, Specificity, and F1 score, and those are calculated by applying the formulas below.

\textbf{Accuracy:} Accuracy is one of the evaluating metrics, and informally interpreted in Equation \ref{equation:accuracy}. It is the proportion of specifically classified fishes and the total number of fishes in the test set.

\begin{equation}
    Accuracy = \frac{\sum_{i}^{N} P_i}{\sum_{i}^{N} \left | Q_i \right |} \times 100\%
    \label{equation:accuracy}
\end{equation}

Where, ${\sum_{i}^{N} P_i}$ is the number of correct predictions, and ${\sum_{i}^{N} \left | Q_i \right |}$ is the total number of predictions.

For binary classification, Accuracy can also be calculated as follows with Equation \ref{equation:accuracy2}.

\begin{equation}
    Accuracy = \frac{TP + TN}{TP + TN + FP + FN} \times 100\%
    \label{equation:accuracy2}
\end{equation}

Where, $TP$ = True Positives, $TN$ = True Negatives, $FP$ = False Positives, and $FN$ = False Negatives.

\textbf{Precision:} The proposition of classified fishes  (TP) and the ground truth (the sum of TP and FP) defines the precision. It calculates the percentage of accurately classified fishes as Equation \ref{equation:precision}.

\begin{equation}
    Precision = \frac{TP}{TP + FP} \times 100\%
    \label{equation:precision}
\end{equation}

\textbf{Recall or Sensitivity:} The ratio of classified fishes (TP) from the ground truth fish (total number of TP and FN) is defined as Equation \ref{equation:recall}.

\begin{equation}
    Recall \; or \; Sensitivity = \frac{TP}{TP + FN} \times 100\%
    \label{equation:recall}
\end{equation}

\textbf{Specificity:} The ratio of TN and the summation of FP and TN determine the specificity with the equation of \ref{equation:specificity}.

\begin{equation}
   Specificity = \frac{TN}{FP + TN} \times 100\%
    \label{equation:specificity}
\end{equation}

\textbf{F1 score (F-measure):} The metric is calculated as the symphonic average of precision and recall \cite{minh2017deep} as the following equation of \ref{equation:F1}.

\begin{equation}
    F1 \, score = \frac{2 * Precision * Recall}{Precision + Recall}
    \label{equation:F1}
\end{equation}

We cannot only rely on the performance evaluation metrics of F1 score and accuracy. A very high cutoff exaggerates the accuracy of a model. So, we also measure the FPR (False Positive Rate), FNR (False Negative Rate), and TPR (True Positive Rate) by the following equation of \ref{equation:fpr} to \ref{equation:tpr}.
\begin{equation}
     FPR  = \frac{FP}{FP + TN} \times 100\%
    \label{equation:fpr}
\end{equation}
\begin{equation}
     FNR  = \frac{FN}{FN + TP} \times 100\%
    \label{equation:fnr}
\end{equation}
\begin{equation}
     TPR  = \frac{TP}{TP + FN} \times 100\%
    \label{equation:tpr}
\end{equation}

We use the Receiver Operating Characteristics or ROC curve for additional evaluation. We can estimate the area under the ROC curve (AUC) with the valuing of the ROC curve \cite{bradley1997use}. The proportion of TPR and FPR from equations \ref{equation:fpr} and \ref{equation:tpr} generates the ROC curve. ROC curve can easily figure out the performance of a model or classifier to differentiate between classes. Higher the AOC means more excellent the classifier or model is predicting. 
\section{Experimental Results}

This section palpates our SVM model results to inspect our model's robustness and see the outcomes of our utilized techniques in both the regular and augmented datasets. Here, we present the actual upshots and comparisons with some graphical representations and tables.

\begin{figure*}
    \centering
     \begin{subfigure}{.24\linewidth}
       \centering\includegraphics[scale=.20]{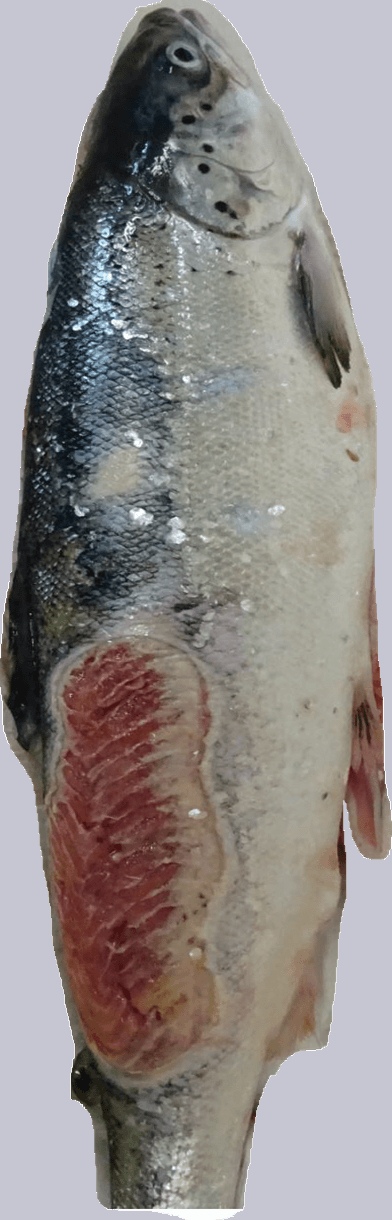}
       
     \end{subfigure}
     \begin{subfigure}{.24\linewidth}
       \centering\includegraphics[scale=.20]{Images/SalmonFish/ImageProcessing/salmon_dis_31.png}
       
     \end{subfigure}
        \begin{subfigure}{.24\linewidth}
       \centering\includegraphics[scale=.20]{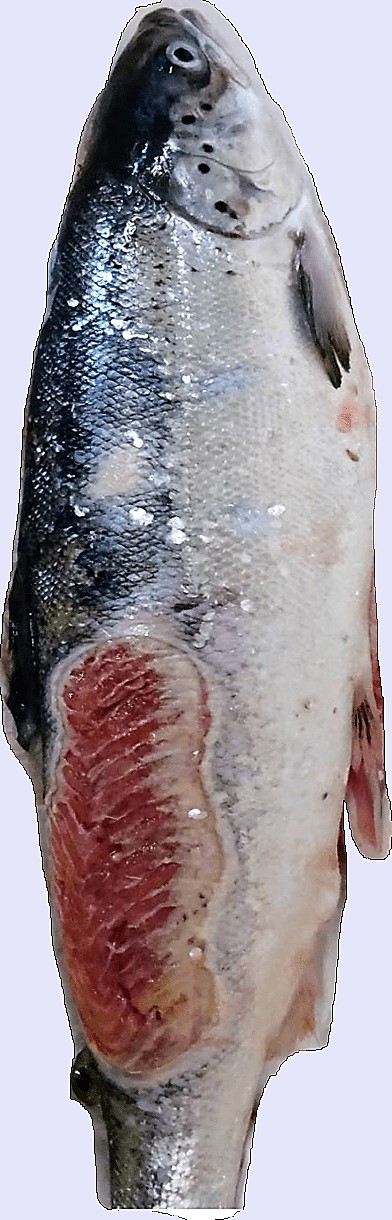}
       
     \end{subfigure}
        \begin{subfigure}{.24\linewidth}
       \centering\includegraphics[scale=.15]{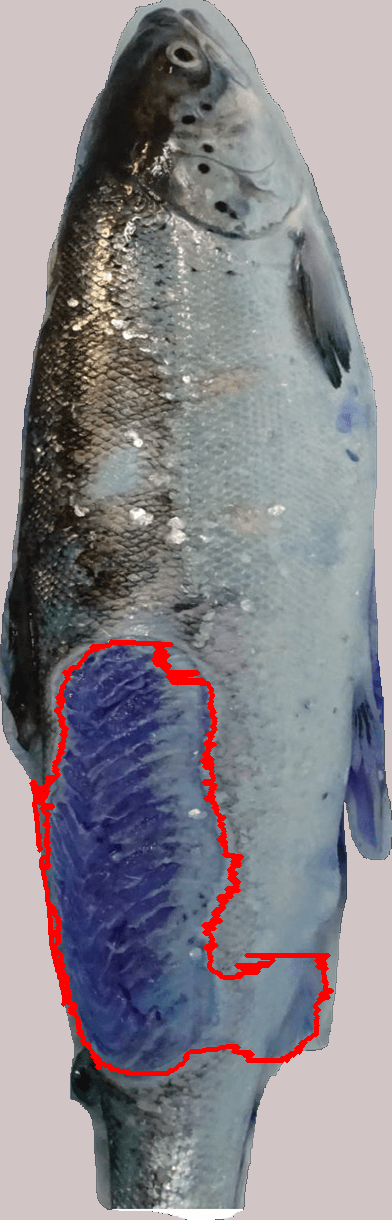}
       
     \end{subfigure}
     
    \medskip
     \begin{subfigure}{.24\linewidth}
       \centering\includegraphics[scale=.20]{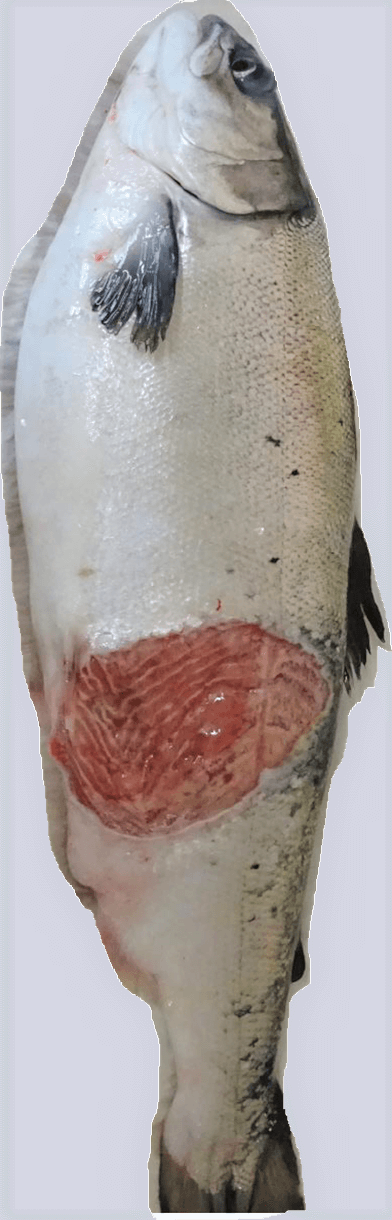}
       \caption{Input image}
     \end{subfigure}
     \begin{subfigure}{.24\linewidth}
       \centering\includegraphics[scale=.20]{Images/SalmonFish/ImageProcessing/salmon_dis_32.png}
       \caption{Resized image}
     \end{subfigure}
        \begin{subfigure}{.24\linewidth}
       \centering\includegraphics[scale=.20]{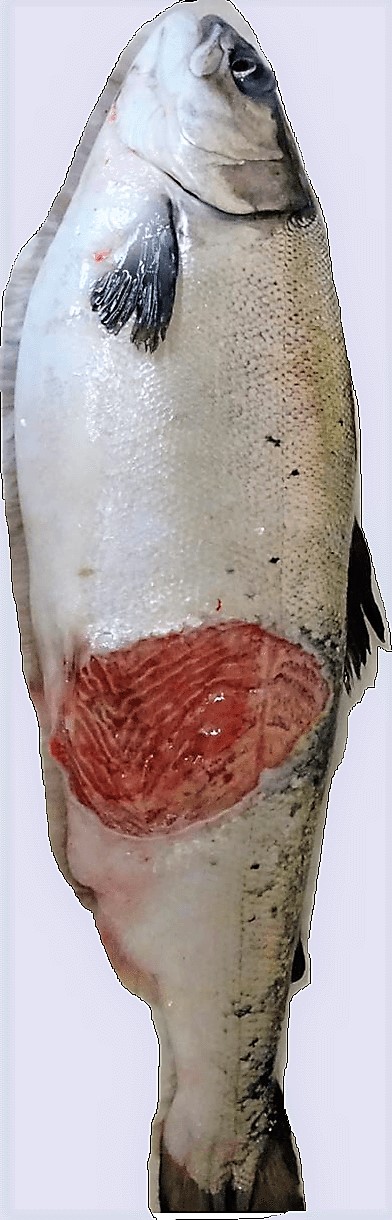}
       \caption{Contrast enhanced image}
     \end{subfigure}
        \begin{subfigure}{.24\linewidth}
       \centering\includegraphics[scale=.15]{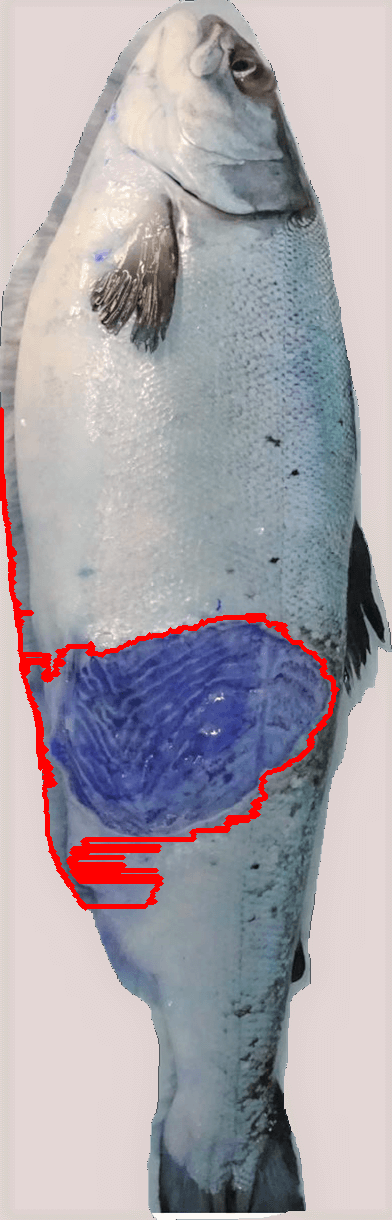}
       \caption{k-means segmented image}
     \end{subfigure}
    
    \medskip

    \caption{Various appearances of image processing (Exhibit the four stages of image processing before features extraction).}
    \label{FIG:salmonStages}
\end{figure*}

First, an input image with any dimension is converted and magnified with a fixed size of 600 $\times$ 250 pixels according to our proposed framework. The image is then segmented into various regions utilizing the \textit{k}-means clustering technique. As a result, a fish image is easily identifiable in terms of the infected and fresh areas. After this segmentation, the infected areas are more observable. All the mentioned aspects are shown in Figure \ref{FIG:salmonStages}.

\subsection{Classification Performance of Proposed SVM}

The classification assessment of our proposed SVM classifier describes in Table \ref{tab:ClassificationResult} (with augmentation) and Table \ref{tab:ClassificationResultWithAugmentation} (without augmentation). Here, both tables only comprise the SVM classifier that concession two classes: fresh fish and infected fish. In Table \ref{tab:ClassificationResult}, for the class of fresh fish, a high percentage shows the sensitivity of 98.46\% whereas the accuracy of 92.0\%. There is also the precision, F1 score, and specificity of 92.75\%, 95.52\%, and 50.00\% correspondingly. In the infected fish class, the highest percentage of 96.02\% shows in the F1 score. Here, the accuracy of 93.50\% and the recall of 98.13\% are also mentioned. Table \ref{tab:ClassificationResultWithAugmentation} shows the accuracy of 93.75\% in the fresh fish class and 94.90\% in the infected fish class. It also shows a good F1 score of 96.23\% and 97.08\% in the fresh and infected fish class.

We distinctly observe Table \ref{tab:ClassificationResult} and Table \ref{tab:ClassificationResultWithAugmentation} where the infected fish class accuracy is 93.50\% and 94.90\%, higher than the fresh fish class. We also see the FPR and FNR in both classes, and the infected class shows slightly higher FNR and lower FPR than the fresh fish class. Here, the low percentage of FPR and FNR refers to our model is not underfit or overfit. So, as an individual class prediction, the infected fish class performs satisfactorily.

\begin{table*}[]
\caption{Class-wise classification results of SVM (Without augmentation).}
\label{tab:ClassificationResult}
\begin{tabular}{ccccccccc}
\hline
\textbf{Classifier}  & \textbf{Class} & \textbf{\begin{tabular}[c]{@{}c@{}}Accuracy\\ (\%)\end{tabular}} & \textbf{\begin{tabular}[c]{@{}c@{}}Precision\\ (\%)\end{tabular}} & \textbf{\begin{tabular}[c]{@{}c@{}}Recall/Sensitivity\\ (\%)\end{tabular}} & \textbf{\begin{tabular}[c]{@{}c@{}}Specificity\\ (\%)\end{tabular}} & \textbf{\begin{tabular}[c]{@{}c@{}}F1-score\\ (\%)\end{tabular}} & \textbf{\begin{tabular}[c]{@{}c@{}}False positive\\ Rate (\%)\end{tabular}} & \textbf{\begin{tabular}[c]{@{}c@{}}False negative\\ Rate (\%)\end{tabular}} \\ \hline
\multirow{2}{*}{SVM} & Fresh fish     & 92.0                                                             & 92.75                                                             & 98.46                                                                      & 50.0                                                                & 95.52                                                            & 50.0                                                                        & 1.54                                                                        \\ \cline{2-9} 
                     & Infected fish  & 93.50                                                            & 94.01                                                             & 98.13                                                                      & 75.0                                                                & 96.02                                                            & 25.0                                                                        & 1.875                                                                       \\ \hline \bottomrule
\end{tabular}
\end{table*}

\begin{table*}[]
\caption{Class-wise classification results of SVM (With augmentation).}
\label{tab:ClassificationResultWithAugmentation}
\begin{tabular}{ccccccccc}
\hline
\textbf{Classifier}  & \textbf{Class} & \textbf{\begin{tabular}[c]{@{}c@{}}Accuracy\\ (\%)\end{tabular}} & \textbf{\begin{tabular}[c]{@{}c@{}}Precision\\ (\%)\end{tabular}} & \textbf{\begin{tabular}[c]{@{}c@{}}Recall/Sensitivity\\ (\%)\end{tabular}} & \textbf{\begin{tabular}[c]{@{}c@{}}Specificity\\ (\%)\end{tabular}} & \textbf{\begin{tabular}[c]{@{}c@{}}F1-score\\ (\%)\end{tabular}} & \textbf{\begin{tabular}[c]{@{}c@{}}False positive\\ Rate (\%)\end{tabular}} & \textbf{\begin{tabular}[c]{@{}c@{}}False negative\\ Rate (\%)\end{tabular}} \\ \hline
\multirow{2}{*}{SVM} & Fresh fish     & 93.75                                                            & 96.23                                                             & 96.23                                                                      & 81.82                                                               & 96.23                                                            & 18.19                                                                       & 3.77                                                                        \\ \cline{2-9} 
                     & Infected fish  & 94.90                                                            & 98.52                                                             & 95.68                                                                      & 88.89                                                               & 97.08                                                            & 11.11                                                                       & 4.31                                                                        \\ \hline \bottomrule
\end{tabular}
\end{table*}

\begin{figure*}[!h]
    \centering
     \begin{subfigure}{.48\linewidth}
       \centering\includegraphics[width=\linewidth]{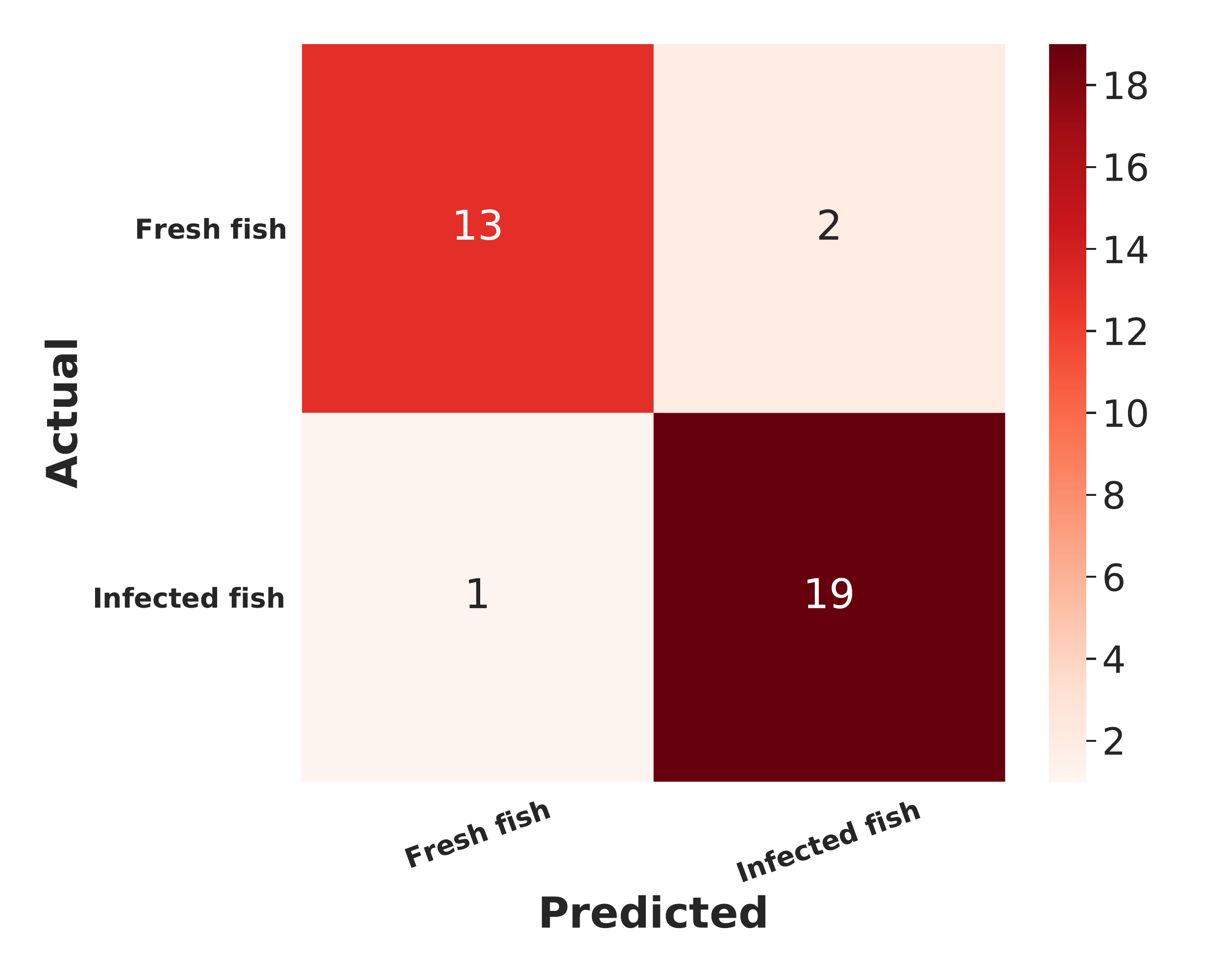}
       \caption{Without augmentation}
     \end{subfigure}
     \begin{subfigure}{.48\linewidth}
       \centering\includegraphics[width=\linewidth]{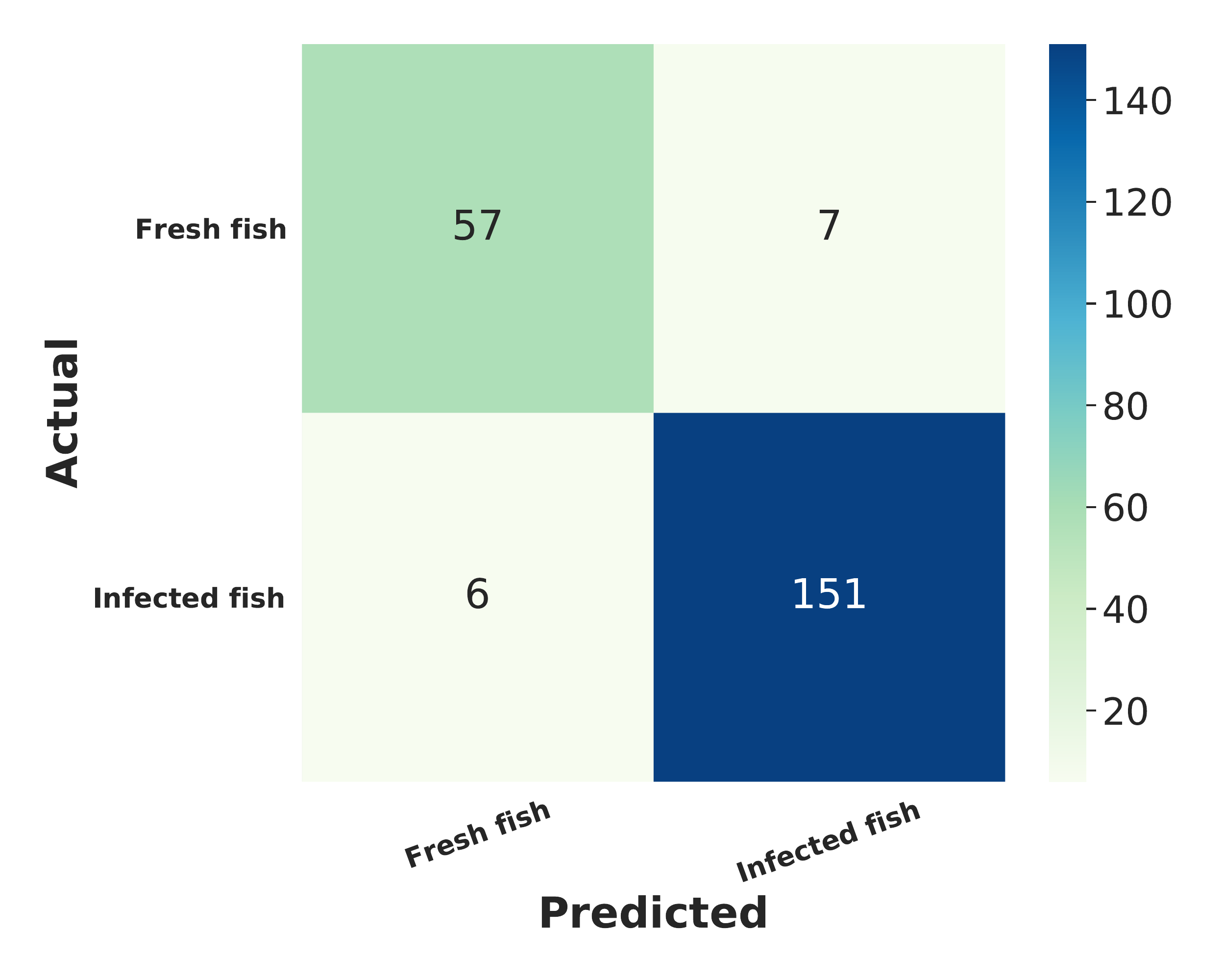}
       \caption{With augmentation}
     \end{subfigure}
    
    \caption{Confusion matrix for SVM classifier.}
    \label{FIG:heatmapSVM}
\end{figure*}

We represent the two confusion matrix as a heat map for better graphical representation shown in Figure \ref{FIG:heatmapSVM} for both with and without augmentation. This heat map can conveniently display the classification and misclassification of our binary class. From this confusion matrix in Figure \ref{FIG:heatmapSVM} (a), we see that fresh fish is misclassified only two times with infected fishes, and infected fish is misclassified once with fresh fish. Figure \ref{FIG:heatmapSVM} (b) shows the misclassification of seven fresh fish with infected fish and six infected fish with fresh fish.


We illustrate our proposed classifier SVM's overall performance in Table \ref{tab:matricEvaluation} to perceive our SVM classifier's metrics-wise performance for both with and without augmentation. From the table, the accuracy is 91.42\% in terms of without augmentation. Furthermore, 94.12\% in terms of augmentation, which is reliable for detecting an infected fish.

\begin{table*}[]
\caption{Metric evaluation of SVM classifier.}
\label{tab:matricEvaluation}
\begin{tabular}{|c|c|c|}
\hline
\multirow{2}{*}{\textbf{Evaluation metric}} & \multicolumn{2}{c|}{\textbf{Value (\% )}}                                                            \\ \cline{2-3} 
                                            & \multicolumn{1}{l|}{\textbf{Without augmentation}} & \multicolumn{1}{l|}{\textbf{With augmentation}} \\ \hline
Accuracy                                    & 91.42                                              & 94.12                                           \\ \hline
Precision                                   & 86.67                                              & 89.06                                           \\ \hline
Recall or Sensitivity                       & 92.86                                              & 90.48                                           \\ \hline
Specificity                                 & 90.48                                              & 95.57                                           \\ \hline
F1-score                                    & 89.66                                              & 89.76                                           \\ \hline
False positive rate                         & 4.43                                               & 9.52                                            \\ \hline
False negative rate                         & 9.52                                               & 7.14                                            \\ \hline 
\end{tabular}
\end{table*}

\begin{figure*}[!h]
    \centering
     \begin{subfigure}{.48\linewidth}
       \centering\includegraphics[width=\linewidth]{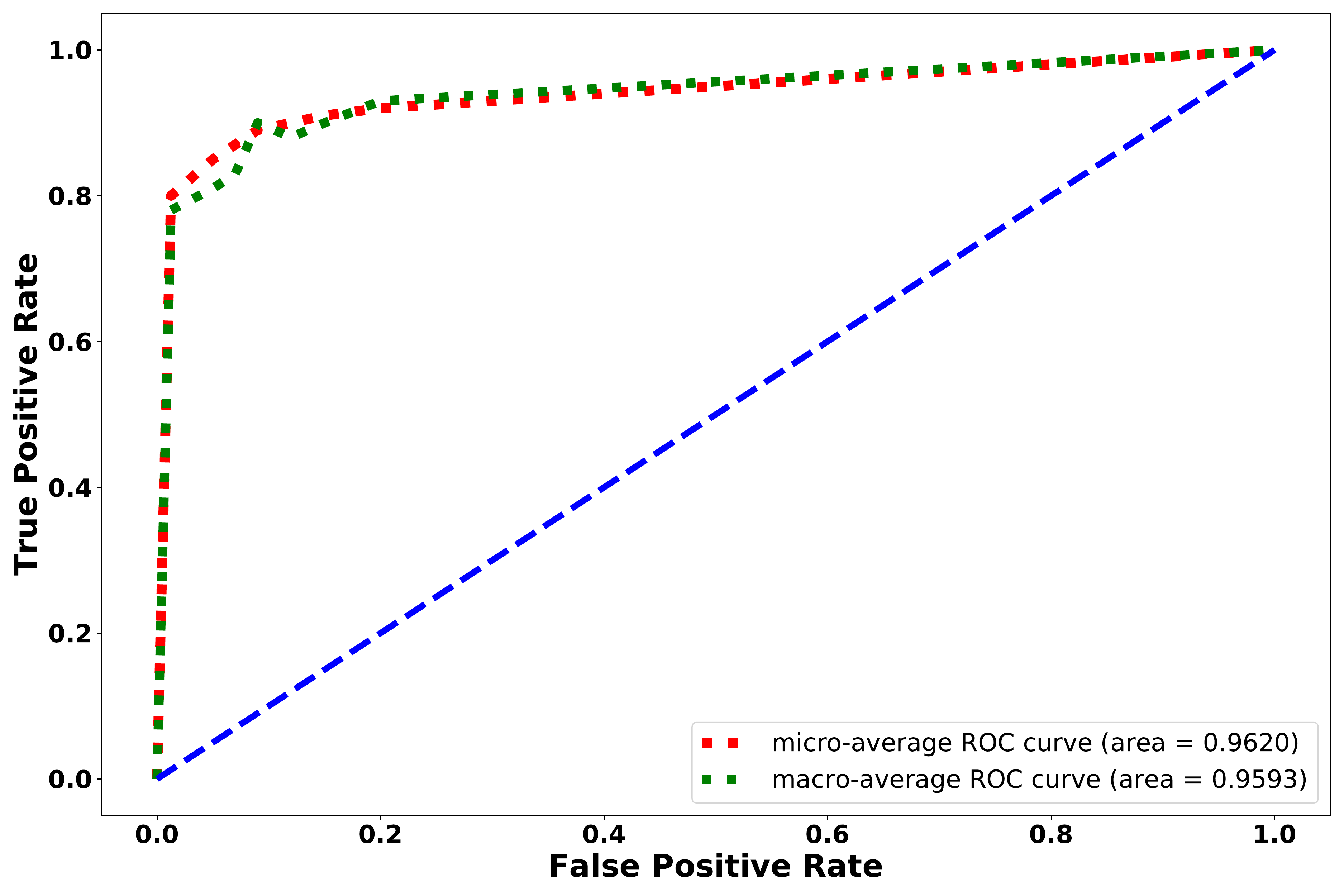}
       \caption{Without augmentation}
     \end{subfigure}
     \begin{subfigure}{.48\linewidth}
       \centering\includegraphics[width=\linewidth]{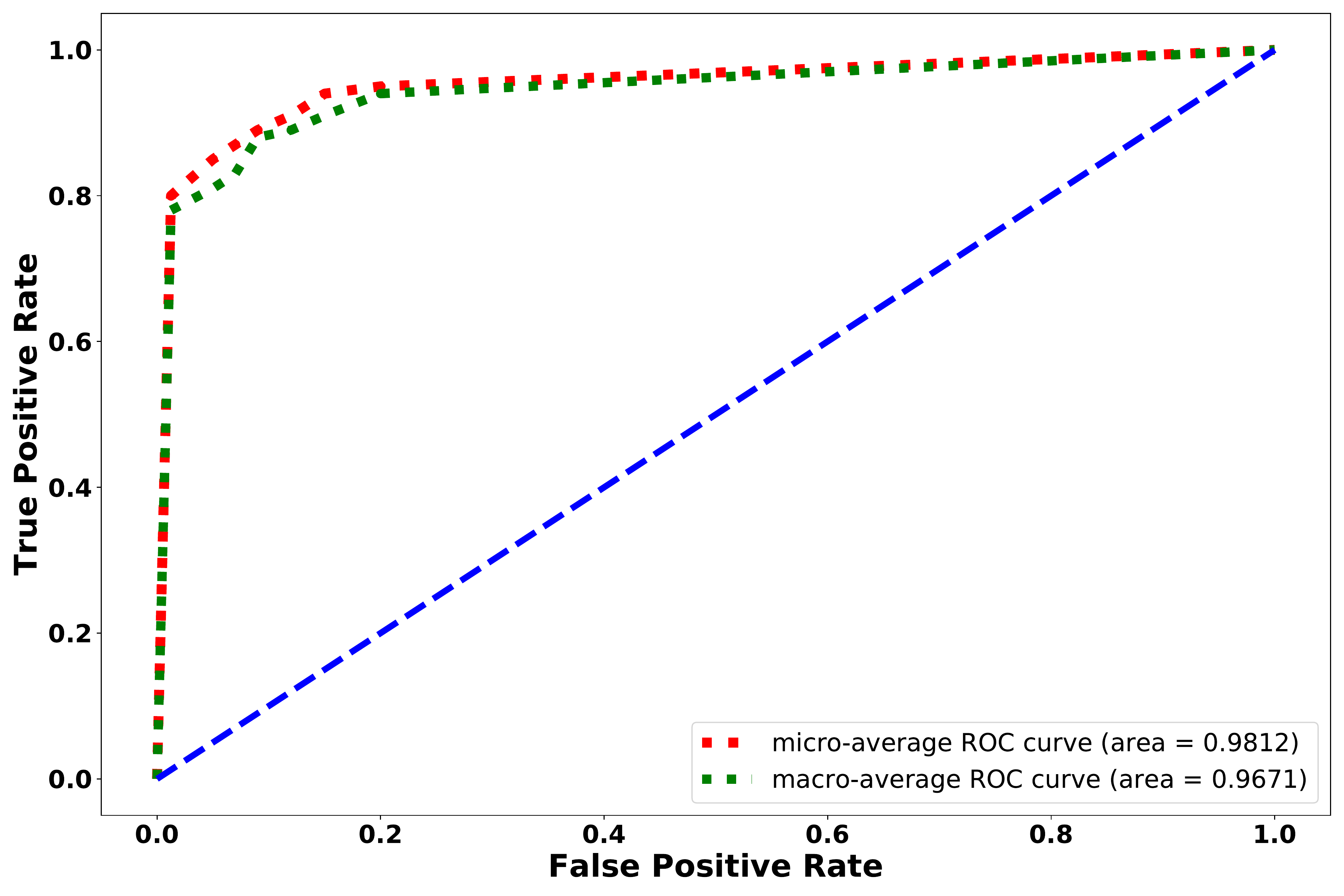}
       \caption{With augmentation}
     \end{subfigure}
    
    \caption{ROC curve for SVM classifier.}
    \label{FIG:rovSVM}
\end{figure*}

In Figure \ref{FIG:rovSVM}, we exhibit a ROC curve that evinces the outcomes to reflect the comprehensive classification performance of SVM with and without augmentation. It scrutinizes the proportion between the true positive rate (TPR) and the false positive rate (FPR). 
Figure \ref{FIG:rovSVM} (a) shows the overall micro average AUC score of 96.20\% and the macro average AUC score of 95.93\% without augmentation. And the Figure \ref{FIG:rovSVM} shows the (b) micro average AUC score of 98.12\% and the macro average AUC score of 96.71\% with augmentation.


\begin{figure*}
    \centering
   
       \centering\includegraphics[scale=.46]{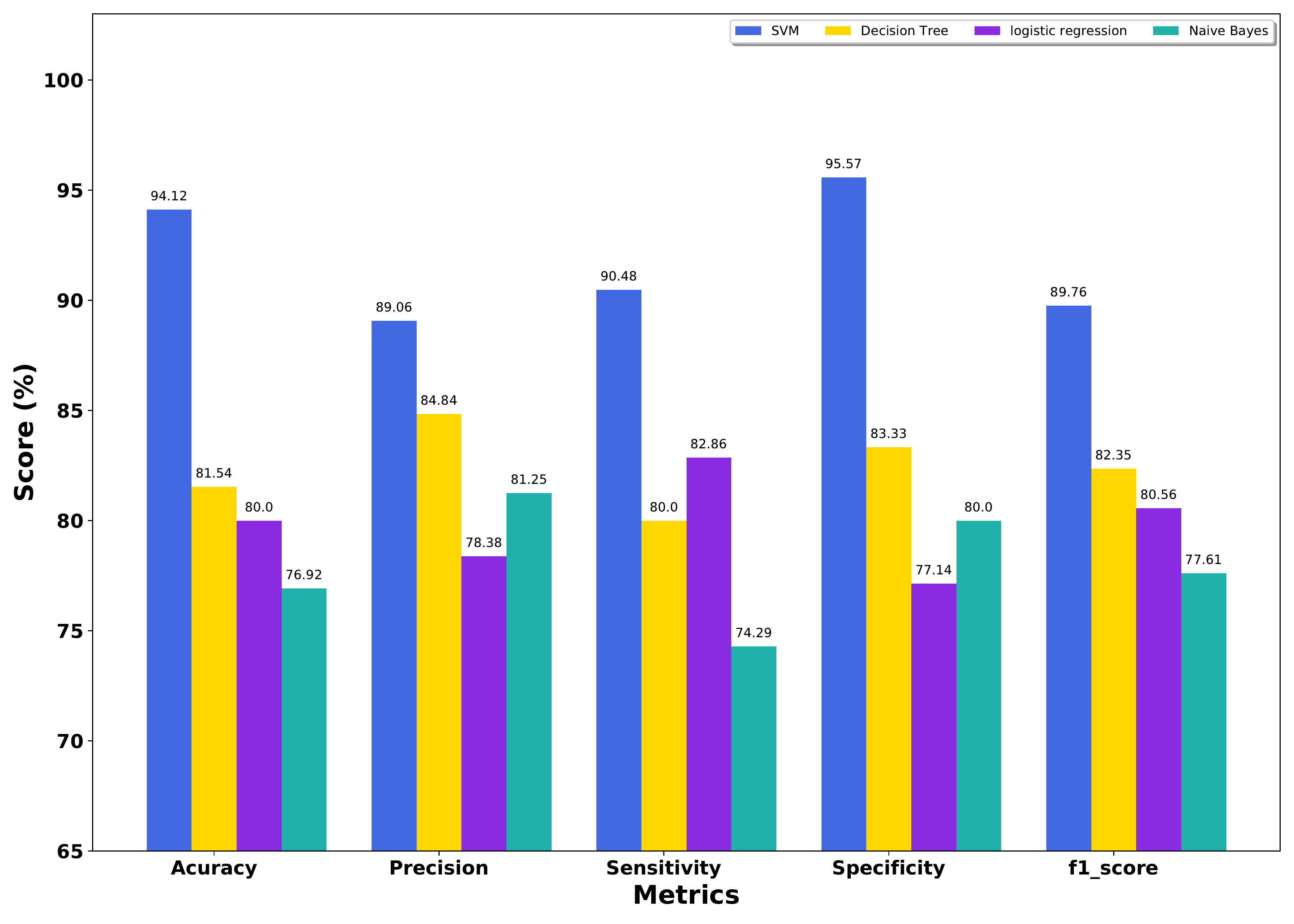}
          \caption{Comparison of classifiers evaluation metrics with image augmentation (Value of Accuracy, Precision, Sensitivity, Specificity and F1-score for SVM, Decision Tree, Logistic Regression and Naive Bayes).}
  \label{FIG:comparisonClassifier}
\end{figure*}

In this paper, we have analyzed SVM so far, but we likewise investigate the other three classifiers for the performance appraisal. Our investigated three classifiers are decision tree, logistic regression,  and naïve Bayes. In Figure \ref{FIG:comparisonClassifier}, we sketch a bar diagram representing the evaluation metrics of our investigated all four classifiers. Here, SVM’s evaluation metrics with augmentation perform more vital than the other three classifiers. Decision tree performs more reliable than logistic regression with an accuracy of 81.54\% whereas logistic regression’s accuracy is 80.0\%, which is better than naïve Bayes. The remaining metrics precision, sensitivity, specificity, and F1-score for the decision tree are 84.84\%, 80.0\%, 83.33\%, and 82.35\% respectively, lower than SVM higher than logistic regression and naïve Bayes.

\begin{figure*}
    \centering
     
     \begin{subfigure}{.45\linewidth}
       \centering\includegraphics[scale=.23]{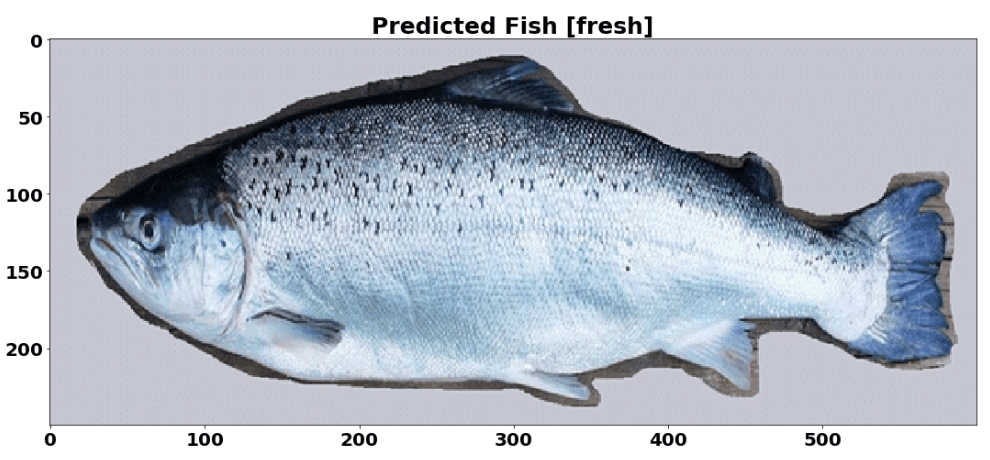}
       \caption{Original: Fresh Fish}
     \end{subfigure}
     \begin{subfigure}{.45\linewidth}
       \centering\includegraphics[scale=.23]{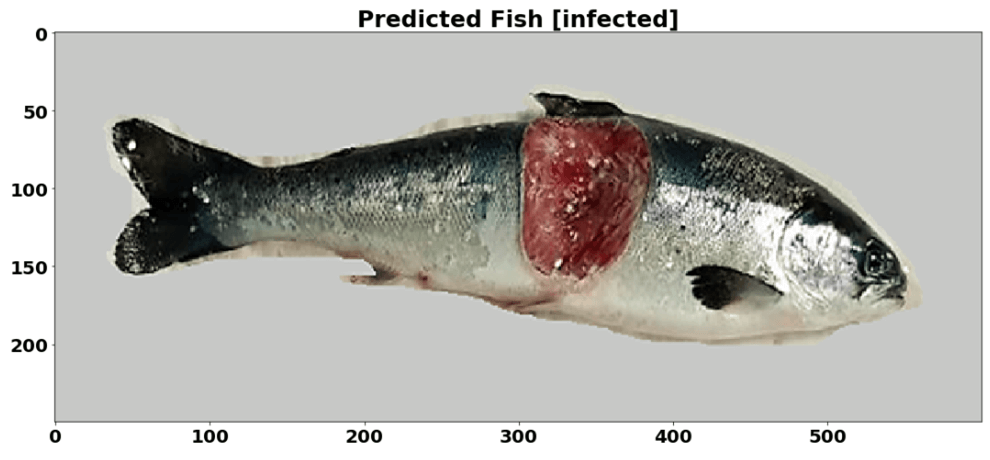}
       \caption{Original: Infected Fish}
     \end{subfigure}
    
    \medskip

    \caption{Fish prediction according to SVM.}
    \label{FIG:predictedFish}
\end{figure*}

Last, we see our predicted result from our SVM classifier in Figure \ref{FIG:predictedFish}. Here, we demonstrate our binary class predictions for fresh fishes and infected fishes. Classifier correctly predicts the inserted original image. 

\subsection{Comparative analysis}

Research works regarding Fish disease detection technique in machine learning mechanisms did not take place up to the mark. The related works are comparatively fewer than other detection work such as fruit disease, crop disease, and some connected work. To appraise our proposed SVM's evaluation metrics to identify the infected fishes, we require studying some relevant promulgated research works. In table \ref{tab:comprisonTable}, we list down some research related to identifying the fish disease. Some works only concentrated on image processing to identify fish disease, and some focused on machine learning based classification models. Shaveta et al. \cite{LRShaveta} arbitrated his work with \textit{k}-means segmentation algorithm and took feature set size of two and then applied neural network classifier to achieve the accuracy of 86\%. Lyubchenko et al. \cite{LRLyubchenko} applied segmentation on this work with the combination of \textit{k}-means clustering and mathematical morphology. This work took three feature sets in image processing and did not apply any classifier. As a result, accuracy is not applicable. Malik et al. \cite{LRMalik} proposed some edge detection method and morphological operation as the segmentation process. This work took three feature sets and applied multiple classification models for comparing the result. Neural network and K-NN (Nearest Neighbour) are the two applied models with 86.0\% and 63.32\% accuracy, respectively.

\begin{table*}[\linewidth,cols=8,pos=h]
\caption{Comparison analysis between this work and related works.}
\label{tab:comprisonTable}
\begin{tabular}{llllll}
\hline
\textbf{Work}                                          & \textbf{\begin{tabular}[c]{@{}l@{}}Segmentation \\ Algorithm\end{tabular}}                                  & \textbf{\begin{tabular}[c]{@{}l@{}}Feature \\ Set\\ Size\end{tabular}} & \textbf{\begin{tabular}[c]{@{}l@{}}Classification\\ Performed\end{tabular}} & \textbf{Classifier}                                                                       & \textbf{\begin{tabular}[c]{@{}l@{}}Accuracy\\ (\%)\end{tabular}}      \\ \hline
This work (with augmentation)                                              & k-means clustering                                                                                          & 10                                                                     & Yes                                                                         & SVM                                                                                       & 94.12 \\ 

This work (without augmentation)                                              &           ---                                                                                          & ---                                                                                                                                             & ---     & ---                                                                                  & 91.42 \\ \hline

Shaveta et al. \cite{LRShaveta}       & k-means clustering                                                                                          & 2                                                                      & Yes                                                                         & Neural network                                                                            & 86.0                                                                  \\ \hline
Lyubchenko et al. \cite{LRLyubchenko} & \begin{tabular}[c]{@{}l@{}}Combination of k-means \\ clustering and \\ mathematical morphology\end{tabular} & 3                                                                      & No                                                                          & Not applicable                                                                            & Not applicable                                                        \\ \hline
Malik et al. \cite{LRMalik}           & \begin{tabular}[c]{@{}l@{}}Edge detection and \\ morphological operation\end{tabular}                       & 3                                                                      & Yes                                                                         & \begin{tabular}[c]{@{}l@{}}Neural network \\ and K-NN (Nearest \\ Neighbour)\end{tabular} & \begin{tabular}[c]{@{}l@{}}86.0 (NN) and \\ 63.32 (K-NN)\end{tabular} \\ \hline
\end{tabular}
\end{table*}


\section{Discussion}

Salmon fish disease detection is an important research area that should require the most attention in the automated research field. However, rarely any intelligent solution for this area comes up in modern times. No existed dataset is available for this research purpose. In this work, we transpire a novel dataset for salmon fish disease detection and conduct our research. Table \ref{tab:datasetSplitting} and Table \ref{tab:ClassificationResultWithAugmentation} have information about our dataset and we divide it for our experiments in this research. Figure \ref{FIG:salmonFishExample} is a little portion of our dataset in which we introduce images of fresh and infected fishes from our dataset. It is mainly defining the input images which we processed and accorded to our classifier.

The main goal we have chased in this research is to classify the infected and fresh salmon fishes. We conduct this experiment based on a real-world image dataset to bring out a reliable system. For ensuring high accuracy, we choose a very efficient machine learning algorithm called support vector machine. SVM is known as one of the leading supervised learning algorithms for classification purposes. In this work, we justify selecting the SVM classifier by comparing our result with other algorithms. The graph from Figure \ref{FIG:comparisonClassifier} justifies our decision to choose the SVM classifier over other techniques. This classifier outperforms to indicate infected and fresh fish for every performance evolution metrics we considered. Comparing Logistic regression, Decision tree, and Naive Bayes, SVM scores higher for accuracy, precision, sensitivity, specificity, and F1 score. These metrics' values from Table  \ref{tab:ClassificationResult} and Table \ref{tab:ClassificationResultWithAugmentation} for our proposed classifier specify the efficiency of this work.

We apply image processing techniques like cubic spline interpolation, adaptive histogram equalization, and k means segmentation before the classification process. In Figure \ref{FIG:salmonStages} we can see how these techniques normalized the raw input image for the classifier.  Figure \ref{FIG:salmonStages}(b) the resized output image of \ref{FIG:salmonStages}(a) is shown. We achieved these resized images after using cubic spline interpolation. Next \ref{FIG:salmonStages}(c) mainly showcasing the contrast-enhanced images resulting from adaptive histogram equalization. This step makes our image dataset more clearer for the classifier. Then we apply \textit{k} means clustering segmentation to differentiate the infected part and fresh part in an image. In \ref{FIG:salmonStages}(d), we display some segmented images from our experiment.

We conduct our experiment by putting the processed images in the proposed SVM classifier. We identify the performance of our classifier by measuring different evolution metrics. It is a good parameter to understand the efficiency of any algorithm for a developed model. In Table \ref{tab:matricEvaluation}, we put the values of these metrics. We sketch a heatmap that describes the confusion metrics for our model. Confusion metrics are a visualization of the performance measurement of any algorithm of machine learning. Figure \ref{FIG:heatmapSVM} (a) and \ref{FIG:heatmapSVM} (b) are the representation of our confusion metrics. Here we can observe the number of misclassification by our classifier is very low.

We present another result in our classifier's justification: the ROC (Receiver Operating Characteristic) curve in Figure \ref{FIG:rovSVM}. This graph mainly conveys our classifier's performance at every possible classification threshold by plotting True Positive Rate and False Positive Rate.

Previously we mention that less research has been conducted for fish diseases, and the existing ones are not up to the mark. Unless this, we find some related works and compare them with our work to make our research more specific. In Table \ref{tab:comprisonTable}, we differentiate our work from other works. The main noticeable difference observed is that none of these works focuses explicitly on salmon fish. One of them uses only image processing techniques which is not an intelligent system. The other two authors have used the neural network but have less accuracy than our classifier. The number of features they considered in their work is lower than ours.

\vspace{-5pt}
\section{Conclusion and Future Work}
We introduce a significant machine learning-based classification model (SVM) to identify infected fishes in this research work. The real-world without augmented dataset (163 infected and 68 fresh) and augmented dataset (785 infected and 320 fresh) are used to train our model is new and novel. We mainly classify fishes into two individual classes: fresh fish and another is infected fish. We appraise our model with various metrics and show the classified outcome with visual interaction from those classification results. Besides developing our classifier, we applied updated image processing techniques like \textit{k}-means segmentation, cubic spline interpolation, and adaptive histogram equalization for transforming our input image more adaptable to our classifier. We also compare our model results with three classification models and observe that our proposed classifier is the best solution in this case.

This work contributes to bringing out a superior automated fish detection system than the existed systems based on image processing or lower accuracy. We not only depend on the modern image processing technique but also adjoin demandable supervised learning techniques. We prosperously develop the classifier that predicts infected fish with the best accuracy rate than other systems for our real-world novel dataset.

In the future, we stratagem to utilize various Convolutional Neural Networks (CNN) architecture for identifying fish disease more precisely and meticulously. Moreover, we will focus on the implementation of a real-life IoT device using the proposed system. Doing so can be a specific solution for the farmers in aquaculture to identify infected salmon fishes and take proper steps before facing any unexpected loss in their farming. We will work with different fish datasets to make our system more usable in other sectors of aquaculture. We will also concentrate on increasing our existing dataset as salmon fish is one of the demanding elements worldwide.

\vspace{-3pt}


\vspace{-5pt}

\bibliographystyle{abbrv}

\bibliography{main}

\begin{thebibliography}{10}

\bibitem{BSPline}
A.~A.~M. Abd~Hamid, N. and A.~Izani.
\newblock Extended cubic b-spline interpolation method applied to linear
  two-point boundary value problem.
\newblock {\em World Academy of Science}, 62, 2010.

\bibitem{acharya2002median}
T.~Acharya.
\newblock Median computation-based integrated color interpolation and color
  space conversion methodology from 8-bit bayer pattern rgb color space to
  24-bit cie xyz color space, 2002.
\newblock US Patent 6,366,692.

\bibitem{agarap2017architecture}
A.~F. Agarap.
\newblock An architecture combining convolutional neural network (cnn) and
  support vector machine (svm) for image classification.
\newblock {\em arXiv preprint arXiv:1712.03541}, 2017.

\bibitem{ben2010user}
A.~Ben-Hur and J.~Weston.
\newblock A user’s guide to support vector machines.
\newblock In {\em Data mining techniques for the life sciences}, pages
  223--239. Springer, 2010.

\bibitem{bianco2007new}
S.~Bianco, F.~Gasparini, A.~Russo, and R.~Schettini.
\newblock A new method for rgb to xyz transformation based on pattern search
  optimization.
\newblock {\em IEEE Transactions on Consumer Electronics}, 53(3):1020--1028,
  2007.

\bibitem{bisong2019google}
E.~Bisong.
\newblock Google colaboratory.
\newblock In {\em Building Machine Learning and Deep Learning Models on Google
  Cloud Platform}, pages 59--64. Springer, 2019.

\bibitem{bradley1997use}
A.~P. Bradley.
\newblock The use of the area under the roc curve in the evaluation of machine
  learning algorithms.
\newblock {\em Pattern recognition}, 30(7):1145--1159, 1997.

\bibitem{burney2014k}
S.~A. Burney and H.~Tariq.
\newblock K-means cluster analysis for image segmentation.
\newblock {\em International Journal of Computer Applications}, 96(4), 2014.

\bibitem{chandra2018survey}
M.~A. Chandra and S.~Bedi.
\newblock Survey on svm and their application in image classification.
\newblock {\em International Journal of Information Technology}, pages 1--11,
  2018.

\bibitem{de2009detection}
L.~de~Oliveira~Martins, G.~B. Junior, A.~C. Silva, A.~C. de~Paiva, and
  M.~Gattass.
\newblock Detection of masses in digital mammograms using k-means and support
  vector machine.
\newblock {\em ELCVIA Electronic Letters on Computer Vision and Image
  Analysis}, 8(2):39--50, 2009.

\bibitem{gaur2015handwritten}
A.~Gaur and S.~Yadav.
\newblock Handwritten hindi character recognition using k-means clustering and
  svm.
\newblock In {\em 2015 4th International Symposium on Emerging Trends and
  Technologies in Libraries and Information Services}, pages 65--70. IEEE,
  2015.

\bibitem{AquacultureIntroduction}
M.~grapple with Atlantic salmon~price rollercoaster.
\newblock Global aquaculture alliance.
\newblock {\em http://www.fao.org/family-farming/detail/en/c/1263890/}, 2020.

\bibitem{hartigan1979algorithm}
J.~A. Hartigan and M.~A. Wong.
\newblock Algorithm as 136: A k-means clustering algorithm.
\newblock {\em Journal of the royal statistical society. series c (applied
  statistics)}, 28(1):100--108, 1979.

\bibitem{ColorCoversion}
H.~Hel-Or.
\newblock Color conversion algorithms.
\newblock
  \url{http://cs.haifa.ac.il/hagit/courses/ist/Lectures/Demos/ColorApplet2/t_convert.html}.
\newblock [Last accessed: 2 Nov, 2020].

\bibitem{hitam2013mixture}
M.~S. Hitam, E.~A. Awalludin, W.~N. J. H.~W. Yussof, and Z.~Bachok.
\newblock Mixture contrast limited adaptive histogram equalization for
  underwater image enhancement.
\newblock In {\em 2013 International conference on computer applications
  technology (ICCAT)}, pages 1--5. IEEE, 2013.

\bibitem{FigAquaculture}
Infected.
\newblock Detecting community structure using label propagation with weighted
  coherent neighborhood propinquity.
\newblock {\em
  https://www.documentcloud.org/documents/4559944-Groatay-Early-2018-2018-0111.html},
  pages 2018--0111, Accessed 10 Jan. 2020.

\bibitem{inoue2017designing}
T.~Inoue, K.~Hasegawa, M.~Yanagisawa, and N.~Togawa.
\newblock Designing hardware trojans and their detection based on a svm-based
  approach.
\newblock In {\em 2017 IEEE 12th International Conference on ASIC (ASICON)},
  pages 811--814. IEEE, 2017.

\bibitem{kailasanathan2001image}
C.~Kailasanathan, R.~S. Naini, et~al.
\newblock Image authentication surviving acceptable modifications using
  statistical measures and k-mean segmentation.
\newblock {\em IEEE-EURASIP Work. Nonlinear Sig. and Image Processing}, 1,
  2001.

\bibitem{khan2016analysis}
S.~Khan, R.~Ullah, A.~Khan, N.~Wahab, M.~Bilal, and M.~Ahmed.
\newblock Analysis of dengue infection based on raman spectroscopy and support
  vector machine (svm).
\newblock {\em Biomedical optics express}, 7(6):2249--2256, 2016.

\bibitem{liu2019adaptive}
C.~Liu, X.~Sui, X.~Kuang, Y.~Liu, G.~Gu, and Q.~Chen.
\newblock Adaptive contrast enhancement for infrared images based on the
  neighborhood conditional histogram.
\newblock {\em Remote Sensing}, 11(11):1381, 2019.

\bibitem{LRLyubchenko}
V.~Lyubchenko, R.~Matarneh, O.~Kobylin, and V.~Lyashenko.
\newblock Digital image processing techniques for detection and diagnosis of
  fish diseases.
\newblock {\em International Journal of Advanced Research in Computer Science
  and Software Engineering}, 6:79--83, 2016.

\bibitem{LRShaveta}
S.~Malik, T.~Kumar, and A.~Sahoo.
\newblock Image processing techniques for identification of fish disease.
\newblock {\em IEEE 2nd International Conference on Signal and Image Processing
  (ICSIP)}, 2017.

\bibitem{LRMalik}
T.~K. Malik, Shaveta and A.~K. Sahoo.
\newblock A novel approach to fish disease diagnostic system based on machine
  learning.
\newblock {\em Advances in Image and Video Processing}, 5.1:49--49, 2017.

\bibitem{marom2010using}
N.~D. Marom, L.~Rokach, and A.~Shmilovici.
\newblock Using the confusion matrix for improving ensemble classifiers.
\newblock In {\em 2010 IEEE 26-th Convention of Electrical and Electronics
  Engineers in Israel}, pages 000555--000559. IEEE, 2010.

\bibitem{meyer2003support}
D.~Meyer, F.~Leisch, and K.~Hornik.
\newblock The support vector machine under test.
\newblock {\em Neurocomputing}, 55(1-2):169--186, 2003.

\bibitem{FishDiseaseIntroduction}
S.~M. Miller and M.~A. Mitchell.
\newblock ornamental fish.
\newblock {\em Manual of Exotic Pet Practice}, Chapter 4:39--72, 2009.

\bibitem{minh2017deep}
D.~H.~T. Minh, D.~Ienco, R.~Gaetano, N.~Lalande, E.~Ndikumana, F.~Osman, and
  P.~Maurel.
\newblock Deep recurrent neural networks for mapping winter vegetation quality
  coverage via multi-temporal sar sentinel-1.
\newblock {\em arXiv preprint arXiv:1708.03694}, 2017.

\bibitem{MariCulture}
M.Phillips.
\newblock Encyclopedia of ocean sciences.
\newblock {\em Academic Press}, pages 537--544, 2008.

\bibitem{FishDiseaseIntroduction2}
B.~L. Nicholson.
\newblock Fish diseases in aquaculture.
\newblock {\em https://thefishsite.com/articles/fish-diseases-in-aquaculture},
  2006.

\bibitem{noble2006support}
W.~S. Noble.
\newblock What is a support vector machine?
\newblock {\em Nature biotechnology}, 24(12):1565--1567, 2006.

\bibitem{rahman2016non}
H.~Rahman, M.~U. Ahmed, and S.~Begum.
\newblock Non-contact heart rate monitoring using lab color space.
\newblock In {\em pHealth}, pages 46--53, 2016.

\bibitem{suthaharan2016support}
S.~Suthaharan.
\newblock Support vector machine.
\newblock In {\em Machine learning models and algorithms for big data
  classification}, pages 207--235. Springer, 2016.

\bibitem{verma2017analysis}
J.~Verma, M.~Nath, P.~Tripathi, and K.~Saini.
\newblock Analysis and identification of kidney stone using k th nearest
  neighbour (knn) and support vector machine (svm) classification techniques.
\newblock {\em Pattern Recognition and Image Analysis}, 27(3):574--580, 2017.

\bibitem{FishFarm}
S.~Winkler.
\newblock How aquaculture works.
\newblock {\em
  https://animals.howstuffworks.com/animal-facts/aquaculture.html}, 19 March
  2020.

\bibitem{zhang2015complete}
J.~Zhang.
\newblock A complete list of kernels used in support vector machines.
\newblock {\em Biochem. Pharmacol.(Los Angel)}, 4:2167--0501, 2015.

\bibitem{zhou2017device}
R.~Zhou, X.~Lu, P.~Zhao, and J.~Chen.
\newblock Device-free presence detection and localization with svm and csi
  fingerprinting.
\newblock {\em IEEE Sensors Journal}, 17(23):7990--7999, 2017.

\end{thebibliography}





\end{document}